\pdfoutput=1
\documentclass[11pt]{article}

\usepackage{ACL2023}
\usepackage{amsmath}
\usepackage{booktabs}
\usepackage{times}
\usepackage{latexsym}
\usepackage{amssymb}
\usepackage{multirow}
\usepackage{hyperref}
\usepackage{url}
\usepackage{booktabs}
\usepackage{graphicx}
\usepackage{caption}
\usepackage{subcaption}

\usepackage{tabularx}

\newcommand{\impro}[1]{{\hspace{0.05cm}{\color[HTML]{32CB00}\textbf{(+#1)}}}}

\newcommand{\weakimpro}[1]{{\hspace{0.05cm}{\color[HTML]{b0eb9d}\textbf{(+#1)}}}}
\usepackage[T1]{fontenc}
\usepackage[utf8]{inputenc}

\usepackage{microtype}

\usepackage{inconsolata}

\newcommand{\cpm}[1]{\textcolor{gray}{$_{\pm #1}$}}

\title{
 Can Language Models Understand Physical Concepts? }
\author{Lei Li\textsuperscript{1}, Jingjing Xu\textsuperscript{2}, Qingxiu Dong\textsuperscript{1}, 
Ce Zheng\textsuperscript{1},  Qi Liu\textsuperscript{3}, Lingpeng Kong\textsuperscript{3}, Xu Sun\textsuperscript{1} \\
   \textsuperscript{1}National Key Laboratory for Multimedia Information Processing, \\
      School of Computer Science, Peking University \\
     \textsuperscript{2}Shanghai AI Lab, 
\textsuperscript{3}The University of Hong Kong \\ 
    \texttt{nlp.lilei@gmail.com} \quad
    \texttt{\{jingjingxu,zce1112zslx,xusun\}@pku.edu.cn} \\
    \texttt{dqx@stu.pku.edu.cn} \quad \texttt{\{liuqi, lpk\}@cs.hku.hk}
  }
\begin{document}
\maketitle
\begin{abstract}

% Vision-language~(V+L) pre-training has shown promising performance in cross-modal tasks such as image-text retrieval and image captioning. On the other hand, these models surprisingly perform worse than text-only models (e.g., BERT) on widely-used text-only understanding tasks. 
% The conflicting results naturally raise a question: 

% How LM understands human world 
Language models~(LMs) gradually become general-purpose interfaces in the interactive and embodied world, where the understanding of physical concepts is an essential prerequisite.
However, it is not yet clear whether LMs can understand physical concepts in the human world. 
To investigate this, we design a benchmark VEC that covers the tasks of (i) \textbf{V}isual concepts, such as the shape and material of objects, and (ii) \textbf{E}mbodied \textbf{C}oncepts, learned from the interaction with the world such as the temperature of objects. 
Our zero (few)-shot prompting results show that
the understanding of certain visual concepts
emerges as scaling up LMs, but there are still
basic concepts to which the scaling law does not apply.
For example, OPT-175B performs close to humans with a zero-shot accuracy of $85$\% on the material concept, yet behaves like random guessing on the mass concept.
Instead, vision-augmented LMs such as CLIP and BLIP achieve a human-level understanding of embodied concepts.
Analysis indicates that the rich semantics in visual representation can serve as a valuable source of embodied knowledge. Inspired by this, we propose a distillation method to transfer embodied knowledge from VLMs to LMs, achieving performance gain comparable with that by scaling up parameters of LMs $134\times$.
\footnote{Our dataset is available at \url{https://github.com/TobiasLee/VEC}}
\end{abstract}

\section{Introduction}

% Intro 
% scaling LMs up solved many tasks, 

% basic 
% After large-scale pre-training, language models~(LMs) have gained impressive achievements in language generation and understanding tasks~\citep{devlin2019bert,radford2019gpt}. 
With the emergent capabilities such as arithmetic~\citep{brown2020language,wei2022emergent} and multi-step reasoning~\citep{chowdhery2022palm} brought by large-scale pre-training, language models~(LMs) are gradually becoming unified interfaces~\citep{hao2022LMasInterface}, capable of instructing embodied robots for high-level tasks such as \emph{cleaning the spilled coke} in interactive and embodied environments~\citep{ahn2022can}. 
% \qi{maybe use another example like grounding LM with robots. There are some recent papers on this topic. Using LM to instruct robots to execute some tasks.} 
Understanding physical concepts is an essential prerequisite for these tasks, e.g., producing correct instructions for cleaning the coke requires understanding the visual characteristics of a coke can, as well as physical properties such as hardness.
However, it still remains unclear whether current LMs can understand basic physical concepts~\citep{driess2023palme}. %, which is crucial for the safe integration of LMs in the embodied scenarios.

% Before utilizing LMs to execute high-level tasks, a sanity check on embodied knowledge such as the temperature of common objects is vital for safely deploying those systems in the human world. 
% Besides, the investigated limitations on those concepts could provide insights for further improvements.
% After scaling up the model parameters and training corpora, LMs are proven to perform better and new capabilities such as arithmetic~\citep{brown2020language} and multi-step reasoning~\citep{chowdhery2022palm} also emerge~\citep{wei2022emergent}.

% Language models~(LMs) have gained impressive achievements in language generation and understanding tasks~\citep{devlin2019bert,Liu2019RoBERTa,radford2019gpt}.
% With further scaling up 

% Scaling law~\citep{kaplan2020scalinglaw} further demonstrates that those abilities could improve along with the increased model scale, computation used and the size training corpora.
% Why we do this Importance of this question: two-fold 
% However, these findings are obtained at a compositional level, leaving the question that whether LMs could understand basic concepts physical in the human world, still unclear. 
% The problem is important 
% as 

% Somehow a little bit weaker
% Second, by examining LMs across different fine-grained perception levels, we could have a better idea of where the current LMs are and make efforts to resolve limitations.

\begin{table*}[t!]
    \centering
    % \scalebox{0.8}{
    \resizebox{\linewidth}{!}{
    \begin{tabular}{@{}l@{\hspace{9pt}}l|l|c|c@{}}
    \toprule
        \multicolumn{2}{c|}{\textbf{Concpet Category}}  & \multicolumn{1}{c|}{\textbf{Instance}} & \textbf{Label} &\textbf{\# of Examples} \\ \midrule

   % & Coref & That, [\textbf{he}]$_1$ says , is just fine with [\textbf{him}]$_2$ . & True  & 27,800\\
   %  & Deps. & [\textbf{Click}]$_2$ [\textbf{here}]$_1$ To view it . & advmod &  25,049\\
   %  \multirow{5}{*}[21pt]{\Wone}  & NER & Back to [\textbf{the Middle East}] tonight . & LOC & 12,586 \\
   %  & SRL& [\textbf{Four Palestinians}]$_2$ were shot and [\textbf{killed}]$_1$ . &  ARG1 & 61,716 \\
   %  & RC & Seniors get much [\textbf{joy}]$_2$ from [\textbf{animals}]$_1$. & Cause-effect & 2,717\\
   %  \midrule 

  &Color & $h$: \textbf{melon}, $t_1$: \textbf{green}, $t_2$: \textbf{black} & green & 574 \\ 
&Shape & $h$: \textbf{lemon}, $t_1$: \textbf{triangle}, $t_2$: \textbf{round} & round & 140 \\ 
\multirow{5}{*}[28pt]{Visual Concepts} &Material & $h$: \textbf{guitar}, $t_1$: \textbf{wood}, $t_2$: \textbf{glass} & wood & 284 \\ 
&Size & $h$: \textbf{ant}, $r$: \textbf{larger than}, $t$: \textbf{bird} & false & 500 \\ 
&Height & $h$: \textbf{bottle}, $r$: \textbf{shorter than}, $t$: \textbf{truck} & true & 500 \\ 
\midrule 
&Mass & $h$: \textbf{wooden spoon}, $r$: \textbf{heavier than}, $t$: \textbf{toaster} & false & 654 \\ 
 \multirow{3}{*}[15pt]{Embodied Concepts} &Temperature & $h$: \textbf{ice}, $r$: \textbf{colder than}, $t$: \textbf{water} & true & 422 \\ 
&Hardness & $h$: \textbf{pearl}, $r$: \textbf{softer than}, $t$: \textbf{glass}& true &  1,016 \\ 
    \bottomrule
    \end{tabular}}
    \caption{The illustration of VEC benchmark.
    % \Wone~focuses on the syntactic or semantic labels of [\textbf{text spans}] and the relations between them. 
    We design two forms of probing tasks. The former~(Color, Shape and Material) asks models to make a choice between two tail options given the head object. The latter~(Size, Height, and all embodied concepts) requires LMs to judge whether the relation is valid given the head and the tail.
    }
    \label{tab:all_dataset_statistics}
\end{table*}

To answer the question, we first define an evaluation suite of physical concepts covering visual and embodied concepts. 
% Specifically, following the definition of \citet{bisk2020experience}, we define evaluation tasks in three world scopes: (1) \emph{the linguistic world}~(\Wone) probing syntactic and semantic knowledge, including tasks like dependency parsing and named entity recognition; (2) 
Specifically, \emph{visual concepts} examine knowledge that can be gained via visual perception, including generic visual concepts, such as color, shape, and material of common objects, and spatial perception, which focuses on the relationship between visual stimuli, i.e., relative size and height of objects.
% ~\citep{liu2022things}
The ability to deal with visual concepts serves as the basis for understanding real-world scenes to perform further instruction.
%  Humans start to develop spatial perception
%and acquire spatial commonsense from infancy, and
% apply the commonsense through lifetime
% visual-related knowledge, including tasks like color-related commonsense understanding and material-related commonsense understanding; 
\emph{Embodied concepts} examine knowledge that requires more interaction and multimodal sensory experience 
in the embodied world, including knowledge about the mass, temperature, and hardness of objects, e.g., ice is colder than water.
% Infants could learn about these concepts of objects in the physical environment by interacting with them~\citep{gopnik1999childrenLearn}.
 % infants learn about objects in their physical environment
% interactive multimodal sensory experience forms the basis of action-oriented
% categories
% Humans gain the corresponding understanding ability from interactive multimodal sensory experience, and 
Understanding embodied concepts is essential for an embodied agent to make correct choices when translating language into actions~\citep{bisk2020experience}.
We compose a Visual and Embodied Concepts evaluation benchmark \textbf{VEC},
with examples shown in Table~\ref{tab:all_dataset_statistics}.
% illustrate .
% in the datasets.
% The tasks in the \textbf{Li}nguistic world, the \textbf{V}isual world, and the \textbf{E}mbodied world together compose our \textbf{LiVE} benchmark.
% \textbf{LiVE}-bench for short. % to comprehensively evaluate the visual-aided language models
% Three research question 

% for the understanding level of physical concepts of different aspects
With the benchmark, we examine popular LMs including text-only LMs and vision-augmented LMs. We cover
masked language models and causal language models in text-only LMs, including BERT~\citep{devlin2019bert} and RoBERTa~\citep{Liu2019RoBERTa}, and GPT~(OPT)-family~\citep{radford2019gpt,zhang2022opt} with parameters ranging from $125$M to $175$B. 
Furthermore, as humans understand the world by learning from multiple modalities, especially using the visual modality~\citep{bloom2002children}, we are interested in whether the vision supervision in recent vision-augmented language models~(VLMs)~\citep{Chen2019UNITER,Radford2021CLIP,wang2021simvlm,madureira-2021-flamingos} could also facilitate the understanding ability of embodied concepts. 
CLIP~\citep{Radford2021CLIP} and BLIP~\citep{li2022blip} are chosen as representatives of VLMs for evaluation, due to their promising performance and the ability to deal with textual-only inputs. 
%impressive results on image representations and cross-modal tasks~\citep{shen2021much} and
% (iii) Most importantly, the LM of CLIP is randomly initialized and pre-trained with a simple contrastive image-text matching objective, which eliminates the effects of parameter initialization and combined pre-training objectives~\citep{Chen2019UNITER}.
% , helping draw clearer conclusions. 
To eliminate the effects of training corpus~\citep{tan2020vokenization},
we re-implement BERT, OPT, and CLIP on the same caption dataset with a similar Transformer model~\citep{vaswani2017attention} for a fair evaluation.  %We further explore different prompting methods to examine the learned knowledge of LMs. 
Previous studies have shown that prompting methods that fit the pre-training paradigm could better elicit the knowledge learned from LMs~\citep{LAMA,timoPET,brown2020language}. We adopt pre-trained-objective style promoting methods to narrow the gap between probing and pre-training.

Our zero (few)-shot results on the VEC benchmark show that: 
% Highlight conclusions 
(i) Moderate-sized LMs such as BERT and RoBERTa exhibit a random-level understanding of both visual and embodied concepts. 
% Further investigation reveals that BERT could produce correct predictions for visual concepts of specific entities, but perform consistently poorly for all embodied knowledge.
% BERT 
(ii) 
A decent visual understanding of specific concepts emerges as LMs scale up, 
% While large causal LMs such as OPT-175B achieve close-to-human performance on visual concepts such as color and material, 
while they still struggle to understand the embodied knowledge with performance slightly better than random guessing.
(iii) Both image-grounded caption text and visual supervision could provide performance gain regarding visual concepts, yet only the latter enhances the understanding of embodied knowledge of LMs. 

We further investigate the source of embodied knowledge in VLMs.
A case study demonstrates that embodied knowledge in the VLM of CLIP is potentially learned from image representations. 
We thus propose a knowledge distillation method to transfer the learned embodied knowledge in VLMs into LMs, resulting in an average accuracy gain of $3.38$, which is comparable to the $4.46$ gain achieved by scaling the model parameters $134$x.
Nevertheless, the improved LMs still exhibit great gaps with humans, indicating great potential for further advancements.
\section{VEC Benchmark}
% In this section, we introduce a comprehensive benchmark for evaluating language models from different perception aspects. %Traditional benchmarks only focus on the linguistic world (e.g., GLUE). Since language models generally have strong linguistic abilities, existing benchmarks are biased toward text-only models. To better evaluate the effects of visual supervision, we argue that a more comprehensive perception evaluation is required.
Our VEC benchmark aims to evaluate the understanding of physical concepts of LMs.
Inspired by the world scope definitions by \citet{bisk2020experience}, we divide physical knowledge into visual knowledge and embodied knowledge.
The former are visual properties that can be acquired via visual perception, while the latter focus on knowledge that requires multimodal sensory interaction.

\subsection{Visual Concepts}
% \qi{add more descriptions about the motivation of using these datasets}
% \xjj{Understanding visual signals is an important aspect of perception modeling.}
Perception is necessary for language learning because it forms the basis for many of our semantic axioms~\citep{bisk2020experience}.
Among the various types of perception, visual concepts model a vastness of experiences in the world that cannot be stated by text alone~\citep{harnad1990symbol}.
% While V+L pre-training has demonstrated great success on downstream cross-modal tasks~\citep{Chen2019UNITER,Li2020OscarOA}, few studies have been conducted towards measuring the visual commonsense knowledge in language models~\citep{zhang-etal-2022-visual,liu-etal-2022-spatial-commonsense}. 
In this work, we consider evaluating the visual understanding ability of LMs by examining their performance on various visual concepts.
%It motivates us to perform a more comprehensive investigation.
Specifically, we combine the recently proposed visual knowledge probing datasets, including Spatial Commonsense~\citep{liu-etal-2022-spatial-commonsense} and ViComTe~\citep{zhang-etal-2022-visual}.
% for evaluation in the visual world
The combined dataset requires not only understanding various generic visual concepts including color, shape, and material, but also understanding the relationship between common objects, such as size and height. % of the common objects
% According to the format of the task definition, 
% These visual-related tasks can be divided into two categories. 
For generic visual concepts, i.e., color, shape, and material identification, we define an answer selection game: selecting a correct value from two options for the attribute given an object. 
For example, given a head object \texttt{banana},
the model should pick the ground-truth tail answer \texttt{yellow} instead of an alternative option such as \texttt{black}.
For visual relationships, i.e., size and height understanding, we define a comparison game: LMs need to perform a comparison between different objects. For example, given a head entity \texttt{ant} and a tail entity \texttt{bird}, the LM is asked to compare the size of two objects and makes a prediction between the correct relation \texttt{smaller} and the false one \texttt{larger}.
% \lei{shall we mention the knowledge conversion here, i.e.,  ant is smaller than table --> QA format?}
%For example, the fact that a coin is usually smaller than a table is represented in \texttt{(ant, smaller than, table)}. The model is asked to compare the size of paired objects by making choices between the ground-truth size description \texttt{smaller than} and the antonym \texttt{larger than}.

\subsection{Embodied Concepts}
% \lei{How to better highlight this part?}

% Interactive
% In addition to learning basic physical properties of the world from interaction
% Understanding the physical realities is also an important aspect of perception. 
The embodied concepts refer to physical realities of objects, e.g., mass, and temperature, 
which infants could learn by interacting with the environment~\citep{gopnik1999childrenLearn}.
This kind of knowledge is the basis of intelligence and enables agent models to explore challenging tasks in physical environments.
% with more challenging tasks for further improving language potential
% and possibly incommunicable by language and images
%Strictly speaking, visual knowledge is also an important part of the physical world. 
We are curious about whether current LMs can capture embodied knowledge via large-scale pre-training. 
In this work, we define embodied knowledge as the knowledge that requires multimodal sensory interaction with the environments beyond visual perception. We construct embodied knowledge evaluation datasets regarding basic physical properties including mass, temperature, and hardness.

\paragraph{Mass Dataset}
We build the Mass dataset by transforming the Image2Mass dataset curated by~\citet{pmlr-v78-image2mass}, which annotates common objects with corresponding weights. 
The most light-weight object in the dataset is a red Lego brick, weighing $0.026$ lbs, and the heaviest object is a $2.664$ lbs drill.
%As directly asking the language model for the absolute mass of objects can be challenging due to the potential poor numeric abilities across scales of language models~\citep{wallace2019numbers},  due to the potential poor numeric abilities across scales of language models
Directly asking the LM for the absolute mass of objects can be challenging~\citep{wallace2019numbers}. We define the task in a comparison format. 
%we turn the problem into a comparison form by asking the language model which object is heavier. for objects with weight annotations in the dataset,
Specifically, each comparison pair contains two objects with a weight gap greater than $1$ lbs.
The threshold is set according to the Weber–Fechner laws~\citep{fechner1948elements} to guarantee that the mass difference is perceivable for humans.
% \footnote{} 
We build $654$ triplets such as \texttt{(hair dryer, heavier than, red Lego brick)} for evaluation.

\paragraph{Temperature Dataset}
We design a temperature probing dataset by collecting the temperature of common objects from Wikipedia.\footnote{\url{https://en.wikipedia.org/wiki/Orders_of_magnitude_(temperature)}}
For example, the ice is $0^{\circ}$C, and the temperature of water vapor is $100^{\circ}$C.
%Consistent with the composition of the mass dataset, w
We convert the object with temperature annotations into pairs, and each pair contains two objects and the corresponding temperature relation. For example, \texttt{(ice, colder than, water vapor)}. The temperature gap between two objects must be greater than a difference threshold, which is loosely set to $10^{\circ}$C for assurance of thermal perception for human~\citep{jones2009thermal}.
The final Temperature dataset consists of $422$ pairs in total.

\paragraph{Hardness Dataset}
% Wiki: 
% In materials science, hardness (antonym: softness) is a measure of the resistance to localized plastic deformation induced by either mechanical indentation or abrasion. In general, different materials differ in their hardness; for example hard metals such as titanium and beryllium are harder than soft metals such as sodium and metallic tin, or wood and common plastics. Macroscopic hardness is generally characterized by strong intermolecular bonds, but the behavior of solid materials under force is complex; therefore, there are different measurements of hardness: scratch hardness, indentation hardness, and rebound hardness.
% Tactile sensing plays a key role in our interaction with the world around us.
Hardness is a measure of the resistance to localized plastic deformation in material science. For example, hard metals such as titanium are harder than soft minerals such as talc.
Humans can perceive the hardness of different materials in interaction with the environment by using tactile organs like fingers~\citep{gueorguiev2016touch}.
To investigate whether LMs capture hardness knowledge, we build a Hardness dataset by collecting the Mohs hardness scores of different objects from Wikipedia.\footnote{\url{https://en.wikipedia.org/wiki/Mohs_scale_of_mineral_hardness}} We define the task in a comparison format. For example, \texttt{(talc, softer than, titanium)}. Each pair contains two objects. The gap between two objects is greater than the threshold for human-level understanding. The final dataset contains $1,016$ pairs.
%The objects are paired with a hardness difference threshold of $0.5$,\lei{this threshold is set arbitrarily, and I did not find some supporting evidence in the literature, should we just say according to our experience?} resulting in the dataset with $1016$ pairs like .

\begin{figure*}
    \centering
    \includegraphics[width=0.85\linewidth]{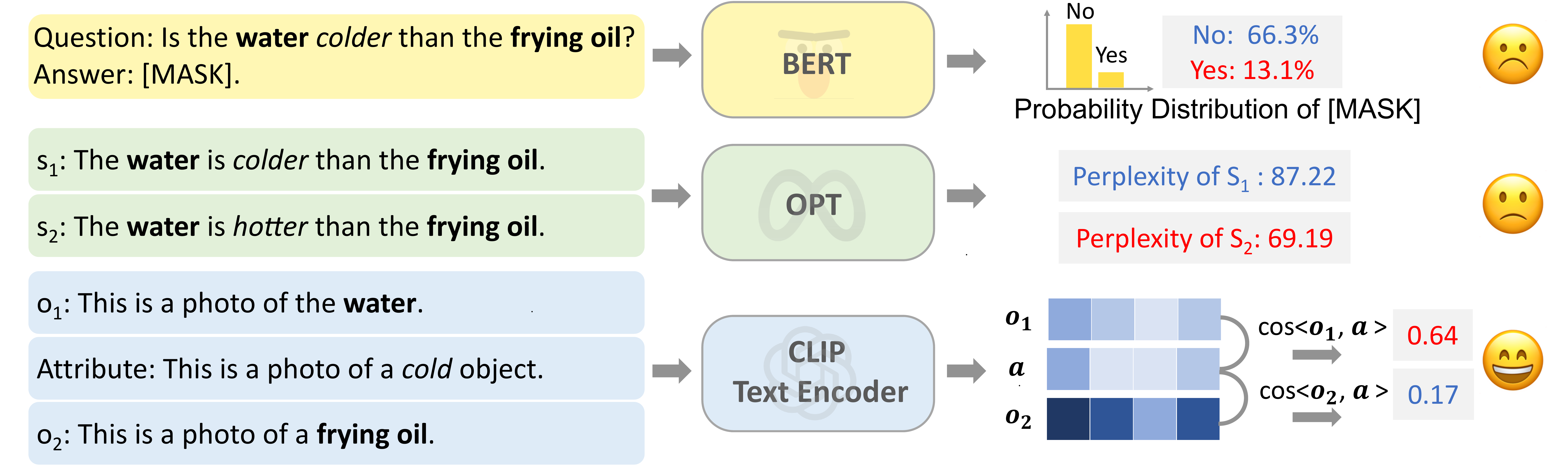}
    \caption{An illustration of prompting methods. For BERT-like models with a masked language head, we convert the knowledge fact to a question and perform prediction with the head over \emph{yes} or \emph{no}. For OPT models, we evaluate the perplexity of different assertions and take the one with lower perplexity as a valid fact. For CLIP, we devise a matching-based probing framework.}
    \label{fig:eval_framework}
\end{figure*}

% The dataset statistics and the illustration examples are shown in Table X.
% \section{ LiVE-bench: Language Model Evaluation}
\section{Prompting Methods}
Recent studies have shown that 
 prompting methods that fit the pre-training paradigm are more effective than other possible prompting methods~\citep{LAMA,timoPET}. Following these studies, we design specific prompts for LMs with different objectives. % we apropose to evaluate the knowledge with multiple manually written prompts.
 %For the VLMs of CLIP without a language model head, we develop a matching-based prompt framework for probing.
 %The details are elaborated below and Figure~\ref{fig:eval_framework} provides an overview.
 % of prompting methods.
 % some 
 % To alleviate the randomness brought by the quality of prompts, we manually write $10$ prompts for each task for an averaged performance, and all the prompts used can be found in Appendix~\ref{apx: prompts}. 

%Specifically, there are two types of knowledge we probed.
% The first type of knowledge is about a specific property of common objects, such as the color of an apple.

\paragraph{Prompting Masked Language Models}
Following PET~\citep{timoPET,schick-schutze-2021-just}, we probe the masked language models by converting knowledge facts into a question-answering form.
For example, a size knowledge fact \texttt{(coin, smaller than, table)} is converted into a sentence with a special mask token: \texttt{Question: is a coin  smaller than a table? Answer: [MASK]}.  We also explored other prompts, such as \texttt{Is a coin [MASK] than table}.
However, our experiments show that a question-answering form can better induce models to generate answers and avoid the influence of tokenization of different LMs. 
Given masked inputs, the model is asked to predict the probabilities of the mask token over two choices, i.e., \texttt{yes} for confirming the knowledge fact is valid or \texttt{no} for an unreasonable assertion.
We observe that in specific LMs, the prediction can be biased toward some answers as investigated by~\citet{Zhao2021CalibrateBU}. 
We calibrate the prediction by normalizing the probabilities according to an estimated prior following \citet{Zhao2021CalibrateBU}. 
%   models towards predicting certain answers
% \qi{this sentence is a bit unclear}
% \lei{is it clearer now?}

\begin{table*}[t!]
\small 
\centering
%\resizebox{0.95\linewidth}{!}{
\begin{tabular}{@{}l|ccccc|c@{}}
\toprule
\textbf{Model~(\# of Param.)} &  \textbf{Color} &  \textbf{Shape} & \textbf{Size} & \textbf{Height} &  \textbf{ Material } & \textbf{Avg.} \\
\midrule

BERT$_\text{YFCC-15M}$ (63M) &	56.05\cpm{10.36} &	53.21\cpm{  1.79}	&50.34\cpm{  1.27}& 	50.16\cpm{ 1.30}  &	55.35\cpm{ 4.79} &	53.02 \\
OPT$_\text{YFCC-15M} $(63M)	& 65.21\cpm{ 15.27} &	51.25\cpm{  18.99} 	&50.50\cpm{  0.77} 	&49.96\cpm{  1.36} 	& 81.41\cpm{
 1.53} 	&59.67 \\ 
CLIP$_\text{YFCC-15M}$ (63M) &	68.21\cpm{  7.17} 	&67.21\cpm{  7.63 }&	62.64\cpm{  6.01} 	&54.04\cpm{  7.05} 	&62.92\cpm{  6.48} &63.00\\ 
\midrule 
BERT-base~(110M)      &  49.29\cpm{  1.60}  &  52.14\cpm{   4.22    }          &  49.94\cpm{  0.80}   & 
 50.56\cpm{  0.59 }&  48.08\cpm{  2.74 }    &     50.00  \\
BERT-large~(340M)     &  49.36\cpm{   1.88 } &   51.21\cpm{ 5.06   }            & 49.26\cpm{   1.60 } & 
  49.08\cpm{ 2.34 }&  49.72\cpm{  0.58 }    &   49.73   \\
RoBERTa-base~(125M)   &  49.07\cpm{  1.62}  &   49.36\cpm{   3.52  }            & 50.32\cpm{  0.57 }  & 
  49.58\cpm{ 0.49 }&  49.86\cpm{  1.44  }   &   49.64   \\
RoBERTa-large~(355M)   &   49.66\cpm{  0.54 } &  50.68\cpm{  1.48    }          &  50.54\cpm{ 1.46  }  & 
 50.14\cpm{  0.45 }&  50.00\cpm{  0.14 }    &   50.20  \\
%	%
\midrule 
OPT~(125M)   &  70.02\cpm{  9.59  } &   57.32\cpm{     6.46  }          &  45.98\cpm{  4.23  }  & 
 56.76\cpm{ 1.36} &  82.43\cpm{  2.20}     &   62.50  \\ 
OPT~(1.3B )  &  76.92\cpm{  5.97 } &   65.00\cpm{     6.12  }         &  51.12\cpm{ 2.66 }   & 
 57.82\cpm{ 4.46 }& 85.63\cpm{ 3.49 }     &     67.30  \\ 
OPT~(13B)   &  79.62\cpm{  5.28 } &  62.50\cpm{   6.44}             & 57.56\cpm{ 6.60 }   & 
 54.58\cpm{  4.53} &  \textbf{88.38\cpm{  3.14}}     &    68.53\\ 
OPT~(175B)   &  \textbf{83.10\cpm{  3.13}}   &   65.71\cpm{   7.54   }          & 59.18\cpm{   9.05  }& 
 55.84\cpm{ 5.33 }&  85.49\cpm{   2.01 }   &    69.87 \\ 

\midrule 

CLIP-ViT/B-32~(63M)  &   80.07\cpm{ 2.57 }  &  84.43\cpm{  2.57    }          & 61.40\cpm{ 6.02}    & 
 62.28\cpm{ 6.40  }    &  80.07\cpm{  2.57}&       73.94   \\
% CLIP-ViT/B-16  &      &   &       &         & &  \\
DeCLIP-ViT/B-32~(63M)  & 81.48\cpm{ 2.63 }&  84.07\cpm{ 2.34}  &        
 \textbf{76.92\cpm{ 1.81}} & 68.12\cpm{ 2.15} &   81.48\cpm{ 2.63} &  78.35 \\
CLIP-ViT/L-14~(123M)  &  80.33\cpm{  3.61}  &   \textbf{85.00\cpm{   4.03} }           &  63.96\cpm{  6.10 }  & 
 60.72\cpm{ 5.56 }&   80.33\cpm{ 3.61  }    &    74.21 \\
BLIP-base~(138M) &    82.60\cpm{  5.50 }&  84.86\cpm{ 2.80  }        &  76.00\cpm{    6.40} & 
\textbf{69.84\cpm{ 7.76}} &  80.67\cpm{ 	1.24 }    &   \textbf{78.79} \\
\bottomrule
\end{tabular}%}
\caption{Zero-shot probing results on visual datasets. Models with the YFCC-15M 
subscript represents that these models are trained from scratch on YFCC-15M data. 
Scaling OPT-family brings clear improvements on size and color datasets. The scaling law fails on the height dataset. 
% Results are averaged over different prompts and the best results are shown in bold.
}
\label{tab:visual_results}
% \vspace{-0.1in}
\end{table*}
\paragraph{Prompting Causal Language Models}
Different from BERT, there is no special \texttt{[MASK]} token in causal language models like GPT~\citep{radford2019gpt}.
Therefore, introducing a special token would result in an inconsistency between pre-training and evaluation.
To remedy this, for each knowledge fact, we state it in natural sentences according to prompting templates and evaluate the sentence perplexity as the proxy metric. 
Specifically, for size-property evaluation, we convert it into a valid knowledge assertion $s1=$ \texttt{A coin is smaller than a table}, and an invalid one by replacing the relation with the antonym adjective $s2=$ \texttt{A coin is larger than a table}. The sentence with lower perplexity is then chosen as the predicted one. %To better extract knowledge from LMs, we manually write diverse prompt templates in this study.
We evaluate the perplexity of each sentence $s=(w_0,w_1,\cdots,w_n)$ as:
\begin{equation*}
\small
% PPL definition 
    \text{PPL}(s) \! = \! P_{\mathcal{M}} (s)^{-\frac{1}{n}} \!= \! \sqrt[n]{ \! \prod_{k=1}^n  \! \frac{1}{P_{\mathcal{M}} \left(w_k \! \mid  \! w_0, \! w_1, \! \ldots,  \! w_{k-1}\right)}}
\end{equation*}
% \begin{align*}
% \small
%     \text{PPL}(s) 
%     &= P_{\mathcal{M}} (s)^{-\frac{1}{n}} \\
%     &= \sqrt[n]{\prod_{k=1}^n \frac{1}{P_{\mathcal{M}} \left(w_k \mid w_0, w_1, \ldots, w_{k-1}\right)}}
% \end{align*}
where $P_\mathcal{M}$ denotes the conditional word probability of the causal language model to be probed and $n$ is the number of tokens in $s$. 
We compare the perplexity $\text{PPL}(s_1)$ and $\text{PPL}(s_2)$ and choose the sentence with lower PPL as a more valid assertion and calculate the prediction accuracy accordingly.
% Need mathematical formulations here

\paragraph{Prompting Vision-augmented Language Models of CLIP}
% It is not feasible to use obtain the probabilities over a pre-defined vocabulary or evaluate the sentence validity via perplexity. 
Unlike masked and causal language models with language head that supports word predictions, the text encoder in CLIP only has one sentence representation without any pre-trained language heads. To probe the learned knowledge in VLMs of CLIP, we design a matching-based prompting method.
In more detail, for the size fact stated before, we first obtain two object descriptions $o_1=$ \texttt{a photo of a coin}, and $o_2=$  \texttt{a photo of a table}.
These two sentences are encoded to get the corresponding object vectors via the CLIP language encoder:
\begin{equation*}
     {\mathbf{o}}_1, {\mathbf{o}}_2 = \text{CLIP}( o_1), \text{CLIP}( o_2). 
\end{equation*}
We then derive an attribute sentence $a=$ \texttt{a photo of a small object}, and encode it to an attribute adjective vector with the language encoder:
\begin{equation*}
    {\mathbf{a}} = \text{CLIP}( a) .
\end{equation*}
% \qi{This paragraph is too vague. You can illustrate the process according to Figure 1.}
The prediction is then performed by comparing the cosine similarity $\text{cos}(\mathbf{o}_1, \mathbf{a})$ and $\text{cos}(\mathbf{o}_2, \mathbf{a})$.\footnote{The matching-based prompting also applies to the pooled embedding of BERT, yet the results exhibit great variance as shown in Appendix~\ref{apx:bert_cls}.} 
The object with higher similarity with the attribute description is adopted as the answer, i.e., a coin is smaller than a table, if $\text{cos}(\mathbf{o}_1, \mathbf{a}) > \text{cos}(\mathbf{o}_2, \mathbf{a})$. Otherwise, we assume that the model thinks the reversed relation holds.
We can also adopt the antonym adjective \emph{large} for getting the attribute vectors. The results of the best-performing adjective words for CLIP are reported and we discuss the influence of adjective options in \S~\ref{subsec:clip_adj}. 

% template.
\section{Experiments}
%In this section, we first introduce the models and prompts used for evaluation, followed by the findings observed on our VEC benchmark.
%Finally, we investigate the mechanism of learned embodied concepts and explore distillation to improve the embodied understanding of LMs.
% \qi{add some sentences describing the contents of each subsection.}
% Does Vision-and-Language Pretraining Improve Lexical Grounding? 
% 用 VisualBERT 训练了一个 captions 版本 在 probing task 上差别不大 initialized with BERT 

\begin{table}[t!]
    \centering
    \footnotesize
    \resizebox{\linewidth}{!}{
\begin{tabular}{@{}l|ccc|c@{}}
\toprule
\textbf{Model~(\# of Param.)} & \textbf{Mass} & \textbf{Temperature} &  \textbf{Hardness} & \textbf{Avg.} \\
\midrule 
BERT$_\text{YFCC-15M}$(63M) &	50.73\cpm{ 2.53 }	&49.50\cpm{ 1.19 }	&50.91\cpm{ 1.04} 	&50.38  \\
GPT$_\text{YFCC-15M}$(63M)	&50.02\cpm{ 0.05} &	57.73\cpm{ 2.24} &	50.04\cpm{ 2.98 }	&52.61  \\ 
CLIP$_\text{YFCC-15M}$(63M) &	67.45\cpm{ 5.16} 	& 64.83\cpm{ 4.17} 	&62.22\cpm{ 3.11  }&	64.83 \\ 
\midrule

BERT-base~(110M)      &  50.35\cpm{ 0.56}  & 49.67\cpm{  0.56 }       &  50.20\cpm{ 0.43 } &      50.07  \\
BERT-large~(340M)     &  49.97\cpm{  1.31 } &  49.83\cpm{  0.50   }              &  49.98\cpm{ 0.06 }  &   49.93  \\
RoBERTa-base~(125M)   &   49.65\cpm{ 0.51 } &  50.00\cpm{ 0.00  }             & 48.04\cpm{ 2.04 }   &  49.23  \\
RoBERTa-large~(355M)  & 50.08\cpm{ 0.23 }   &  50.07\cpm{ 0.19 }    &  49.95\cpm{  0.15}    &  50.03 \\

\midrule
OPT~(125M)  & 50.00\cpm{ 0.00 }&  54.53\cpm{ 4.33} & 46.16\cpm{  2.45}  &  50.23 \\ 
OPT~(1.3B)  & 50.05\cpm{ 0.10 }  & 50.90\cpm{  5.08}  &  53.03\cpm{ 2.69 } &  51.33     \\ 
OPT~(13B) &   50.14\cpm{ 0.36 }  & 51.85\cpm{  6.34} &  52.38\cpm{  3.09 }&  51.46  \\ 
OPT~(175B)  &   50.21\cpm{ 0.24}  &  59.83\cpm{  8.68  }&  57.33\cpm{ 3.41}  &  55.79    \\ 

\midrule

CLIP-ViT/B-32~(63M)  & 65.20\cpm{ 4.75 }   &  60.28\cpm{ 6.83}    &  59.43\cpm{ 2.00 }  &    61.64  \\
% CLIP-ViT/B-16  &      &   &       &         & &  %	%
DeCLIP-ViT/B-32~(63M)  &  54.95\cpm{ 2.00} & 68.58\cpm{ 3.08} &   61.10\cpm{ 4.14}	&   61.54	  \\
CLIP-ViT/L-14~(123M)  &  73.15\cpm{ 6.34 } &    65.88\cpm{	2.31  }  &  \textbf{69.57\cpm{	2.26} }&  69.53 \\
 BLIP-base~(138M)  & \textbf{83.94\cpm{ 2.59} } &   \textbf{74.98\cpm{ 5.60} }    &   56.93\cpm{ 	5.56 }  & \textbf{71.95}  \\
\bottomrule
\end{tabular}}
  \caption{Zero-shot results on embodied datasets. 
    LMs struggle to understand embodied knowledge, including OPT (175B) and visual-augmented LMs, with 71.95 as the best average performance.
  % We report the average performance associated with standard deviation over multiple prompts. The best results are shown in bold.
  }
  % \vspace{-0.18in}
\label{tab:embodied_ret}

\end{table}

\subsection{Experimental Settings}
\label{subsec:exp_setting}

% The applications on BERT~\citep{tenney2019bertpipeline} have shown that the BERT learns a classical NLP pipeline, where the lower layers handle the superficial-level syntactic features. 
% In comparison, the higher layers deal with the semantics.
\paragraph{Models} 
We cover two kinds of LMs, text-only LMs and visual-augmented LMs. 
Text-only LMs include BERT-base~(large)~\citep{devlin2019bert}, RoBERTa-base~(large)~\citep{Liu2019RoBERTa} for masked language models, and OPT models with parameters ranging from $125$M to $175$B.
For VLMs, we include the text encoders of CLIP-ViT-B/32 and CLIP-ViT-L/14~\citep{Radford2021CLIP} as a base and a large version, respectively.
 We also include an enhanced VLM with masked language modeling as self-supervision, DeCLIP-ViT-B/32~\citep{declip} and BLIP, a boosted VLM by unifying multi-modal understanding and generation tasks~\citep{li2022blip}.\footnote{\url{https://huggingface.co/Salesforce/blip-itm-base-coco}}
Since directly comparing the VLMs and text-only LMs can be unfair due to the difference in model configuration and training corpus~\citep{tan2020vokenization}, we re-train CLIP, BERT, and GPT from scratch with a similar Transformer model on the same text corpus, the caption dataset in the YFCC-15M dataset~\citep{thomee2016yfcc100m}. All models are trained for $32$ epochs.
The only difference between these models is the pre-training objective. 
Detailed model and training settings are elaborated in Appendix~\ref{apx: model_parameters}.
% We further add two variants of CLIP, i.e., DeCLIP~\citep{declip} and DeFILIP~\citep{filip}, to investigate the effect of extra language modeling supervision and fine-grained matching, respectively.

% In $\mathcal{W}_2$ and $\mathcal{W}_3$, we additionally evaluate the original models for investigating the learned knowledge during large-scale pre-training.

\paragraph{Prompts} 
% For each task in \Wtwo ~and \Wthree, 
% Prompting methods have been reported to
We manually write several prompts~(at least $4$ prompts for each task) to eliminate the side-effect of the expression variations and report the averaged accuracy. 
Besides, the variance across different prompts could also serve as an indicator to evaluate the robustness of learned knowledge facts. 
% For OPT models, we observe its high sensitivity to the prompts, where some prompts result in significantly worse performance than random guessing. 
We report the averaged performance over multiple prompts for all models.
All used prompts can be found in Appendix~\ref{apx:prompts}.

% \begin{figure}[t!]
%      \centering
%      \includegraphics[width=0.58\linewidth]{iclr2023/plots/radar_max_mdl.pdf}
% \caption{ 
% Linguistic probing results. Higher scores represent richer linguistic knowledge. 
% % Radar chart of the maximum compression ratio of different models in all layers for l
% }
%  \label{fig:w1_probing}
% \end{figure}

\subsection{Main Findings}

\paragraph{The ability of certain visual concepts emerges as scaling up LMs, but there are still basic visual concepts where the scaling law fails.}
The evaluation results on visual datasets are shown in Table~\ref{tab:visual_results}.
% We observe that masked language models such as BERT, RoBERTa achieve a random-guessing level performance on all the visual concepts.
Interestingly, with the scaling up of OPT-family models, the prediction accuracy increases obviously on specific visual concepts such as color and size.
On material and color, the largest OPT-175B model even achieves better results than VLMs of CLIP-ViT/L-14, which are augmented with vision supervision and are supposed to perform better~\citep{zhang-etal-2022-visual,liu2022things}.
%The performance boost with increased model size indicates that the scaling law applies to certain visual concepts.
It is because the combination of color and material frequently occurs (e.g., red apples) in raw texts. The co-occurrence statistics of color and material are well captured by large LMs. This is supported by the performance improvements after training on visual-grounded text corpus YFCC-15M.
% VLMs achieve the best average performance. 
% We also observe that DeCLIP and a larger VLMs of CLIP-ViT/L-14 both achieve better results, indicating that adding extra self-supervision to VLMs and scaling up the model size are both beneficial.
% Most surprisingly, when the OPT LM scales to up $175$B, it can even outperform VLMs on visual properties like color and material, indicating that specific visual-related knowledge can also be acquired via purely language modeling.
However, we also notice that increasing LMs to 175B brings negligible improvements in the Height dataset, indicating that there still remain visual concepts where the scaling law does not hold even though these concepts can be easily captured by humans.

\begin{table}[t!]
    \centering
    \footnotesize
    \resizebox{\linewidth}{!}{
\begin{tabular}{@{}l|ccc|c@{}}
\toprule
\textbf{Model} & \textbf{Mass} & \textbf{Temperature} &  \textbf{Hardness} & \textbf{Avg.} \\

\midrule
Zero-shot Best VLMs & \textbf{83.94\cpm{ 2.59} } &   \textbf{74.98\cpm{ 5.60} }   &  \textbf{69.57\cpm{	2.26 }}&  \textbf{76.16} \\ 
\midrule
BERT-base     &  64.72\cpm{ 4.77} & 55.62\cpm{ 1.34 }      &  51.80\cpm{ 1.31  }&     57.38  \\
BERT-large     &  65.47\cpm{  4.86 } &  54.19\cpm{  2.31  }     &  52.73\cpm{1.22}  &   57.46  \\
RoBERTa-base   &  60.24\cpm{ 5.24}&  	60.27\cpm{ 4.85  }             & 50.44\cpm{ 1.13}    &  56.98 \\
RoBERTa-large  & 61.18\cpm{ 4.01}   &  58.28\cpm{ 2.09  }  & 50.56\cpm{  1.14}  &  56.67 \\

\bottomrule
\end{tabular}}
  \caption{The few-shot results of BERT variants.  With $16$ instances, the fine-tuned BERT variants are still worse than zero-shot visual-augmented LMs.
  % We report the average performance associated with standard deviation over multiple prompts. The best results are shown in bold.
  }
\label{tab:fsk_mlm}
\end{table}

\begin{figure}[t!]
    \centering
    \includegraphics[width=0.8\linewidth]{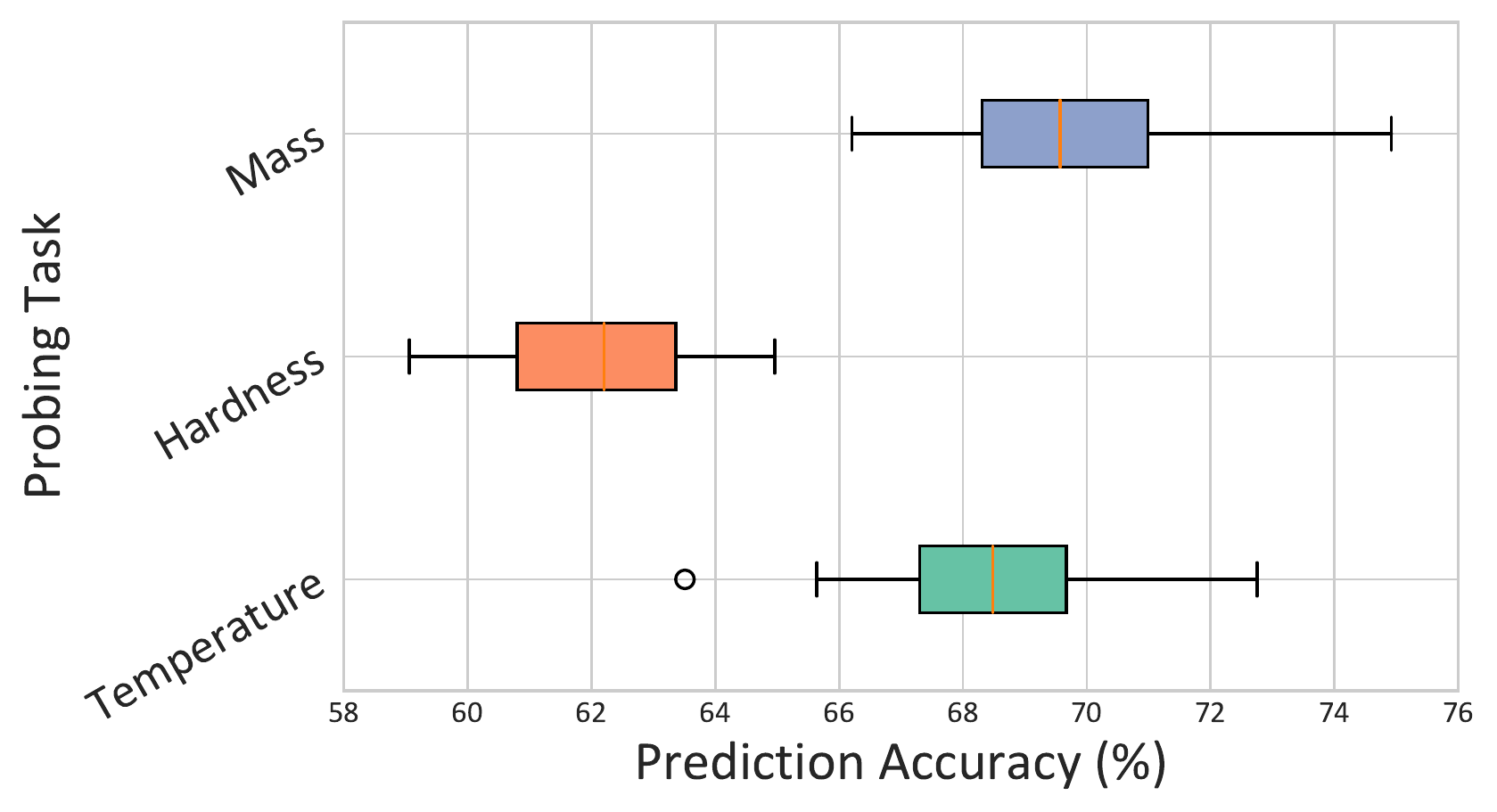}
    \caption{Few-shot results of OPT-175B with $16$ instances as demonstration on embodied tasks.
    }
    \label{fig:fsk_opt}
\end{figure}

\paragraph{LMs exhibit a poor understanding of embodied concepts.}
% We compare the models on the embodied benchmark. 
As shown in Table~\ref{tab:embodied_ret}, the scaling law fails again on the embodied concepts, as all LMs, including OPT-175B and variants trained with captions data, perform poorly.
Although VLMs achieve better results than text-only LMs, they still show poor results, slightly better than random guessing on all three embodied tasks.
% Besides, different from the observations with visual , scaling up the model~(CLIP-ViT/L-14) instead of adding extra language-size supervision~(DeCLIP) is more effective for improving performance in the embodied world.
We further conduct a few-shot prompt evaluation for OPT models by constructing the inputs with $k=16$ randomly sampled instances and adopt PET~\citep{timoPET} for masked language models.
The results are illustrated in Figure~\ref{fig:fsk_opt} and Table~\ref{tab:fsk_mlm}, respectively.
We find that while the performance is boosted, the average results are still worse than the best-performing VLM model without any demonstration, which only utilizes $0.08\%$ parameters of OPT-175B.
These findings show that visual supervision can help learn embodied knowledge, but there is still a large gap between the best results of existing LMs with human performance. % beneficial for learning knowledge in the embodied world, and its role is irreplaceable by massive text data and billion-level parameters.

\begin{figure}[t!]
    \centering
    \includegraphics[width=0.9\linewidth]{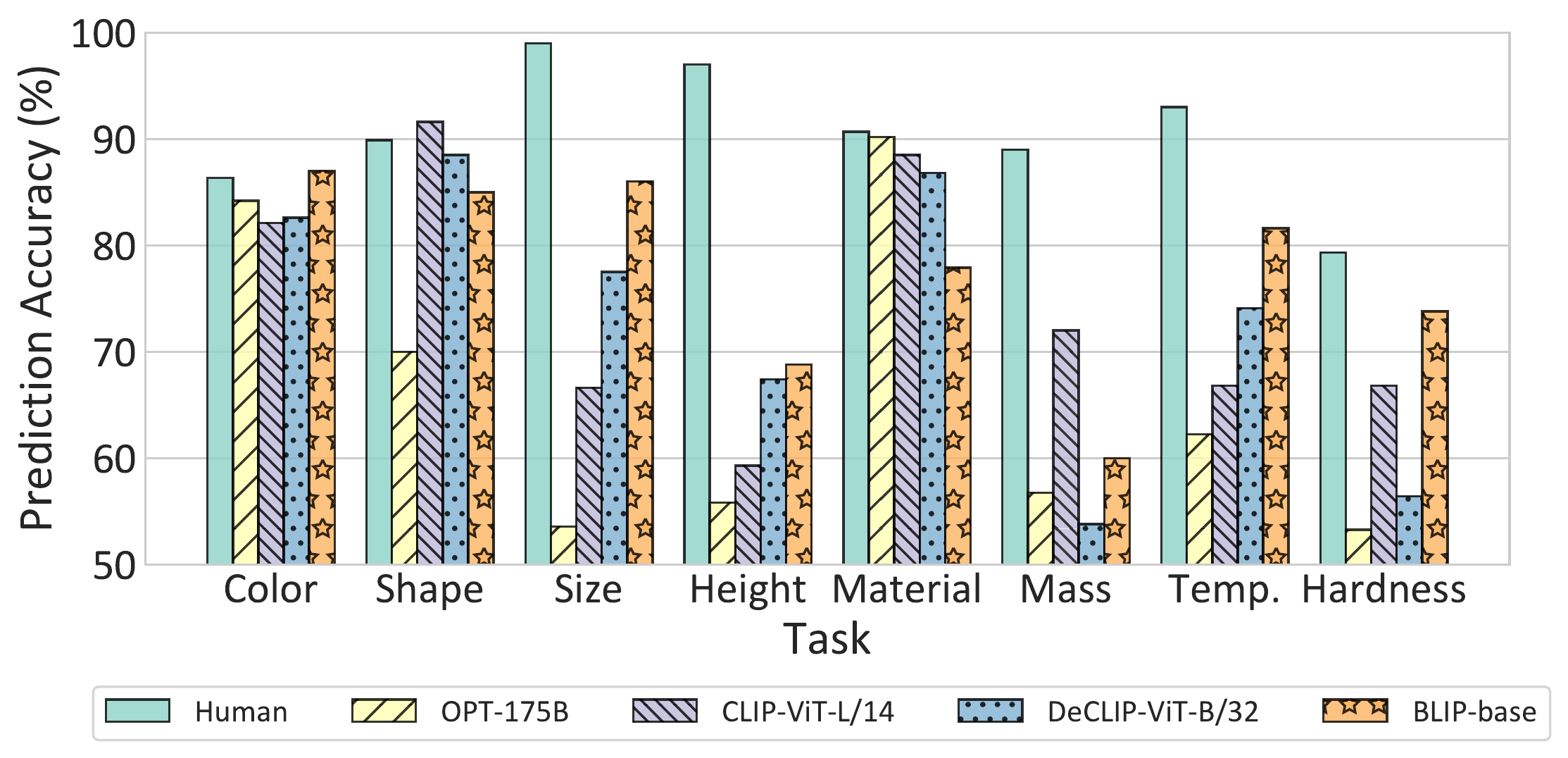}
    \caption{Comparison between the best-performing models and human annotators on sampled subsets of VEC. The best-performing LMs and VLMs achieve close-to-human results on visual datasets, yet far lag behind humans in embodied datasets.}
    \label{fig:human_perform}
\end{figure}

\paragraph{Compared with human annotators, OPT-175B and VLMs achieve competitive performance regarding visual concepts, yet exhibit great gaps with humans on embodied concepts.}
We conduct a human evaluation to better understand the performance of different models.
Specifically, we randomly sample $100$ examples for each task and ask three volunteers to label these examples.
The annotators achieve substantial agreements on all the tasks with Cohen’s kappa~\citep{cohen1960coefficient} $\kappa$ larger than $0.7$, except for the Hardness dataset with a moderate $\kappa=0.52$.
The comparison with best-performing models, i.e., OPT-175B, CLIP-ViT/L-14 and DeCLIP is illustrated in Figure~\ref{fig:human_perform}.
We find that (i) Regarding visual concepts, both OPT and CLIP-like models perform closely to human annotators. CLIP and DeCLIP even outperform the human annotators on the shape task, which is potentially due to the noise introduced by the automatic construction of the dataset~\citep{zhang-etal-2022-visual}.
Overall, the close-to-human results indicate that visual knowledge can be effectively acquired by large-scale cross-modal pre-training or even text-only pre-training with sufficient parameters.
(ii) Regarding embodied concepts, the best-performing CLIP-ViT-L/14 model still has an absolute $18.5\%$ accuracy gap with the humans.
The clear performance gaps reveal that there is still a long way to go in equipping LMs and VLMs with embodied knowledge.

\subsection{Analysis}
\label{subsec:analysis}

\begin{figure}[t!]
    \centering
    \includegraphics[width=0.9\linewidth]{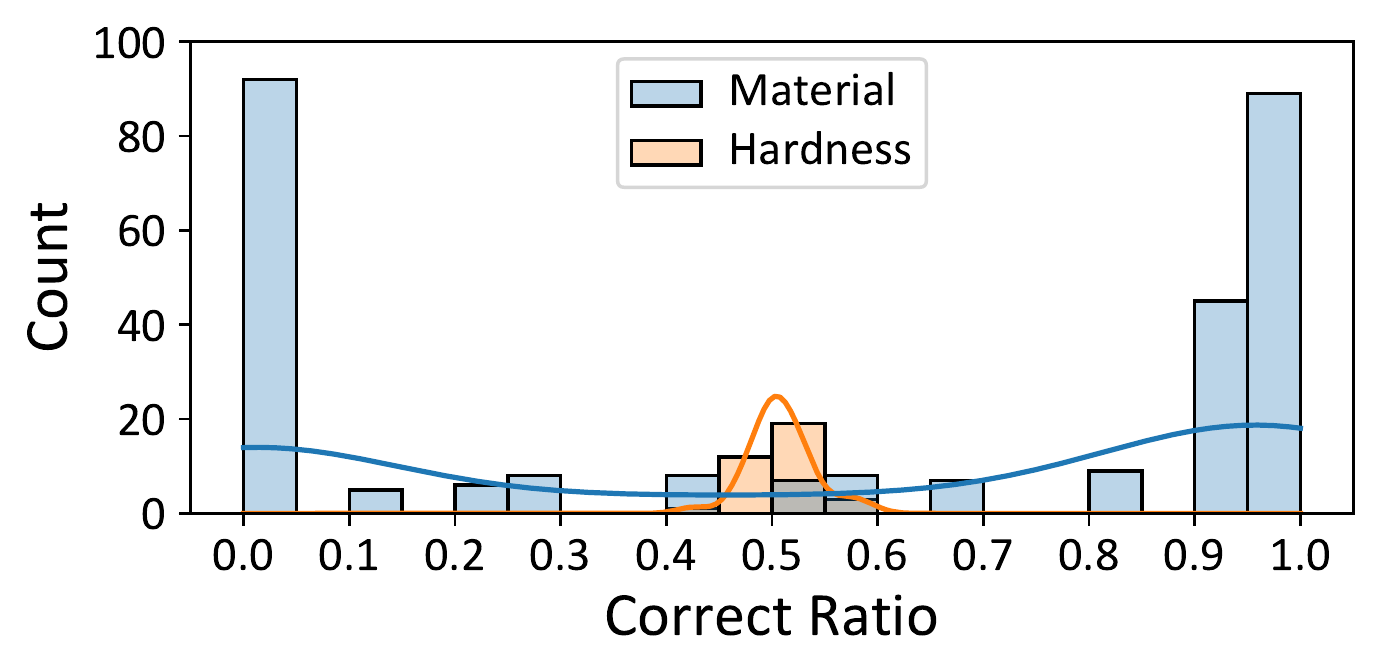}
    \caption{Entity correct ratio histograms of Mass and Material datasets across different prompts. BERT could make consistent correct predictions for specific entities, and the bell curve on the hardness indicates it is challenging for BERT to understand embodied concepts.}
    \label{fig:hist_combine}
\end{figure}

\paragraph{Does BERT behave similarly regarding visual and embodied concepts?}
The overall prediction results of BERT-like models in the visual and embodied world are both at a random level.
We investigate this question result by first checking whether BERT models perform consistently at a guessing level for all the entities in the dataset.
We compute the entity correct ratio among different prompts for the objects in different datasets and compare the distribution on different tasks with the BERT model trained on YFCC-15M dataset. 
As illustrated in Figure~\ref{fig:hist_combine}, in the Material identification task, 
there are entities that the model that could provide consistent correct predictions. 
However, the distribution of the Hardness dataset in embodied evaluation exhibits a bell curve, i.e., most entities are predicted correctly at a random-chance level. The distributions of other tasks show similar results and can be found in Appendix~\ref{apx:entity_dist}.
These results suggest that BERT learns visual knowledge for certain entities yet indeed struggles regarding embodied concepts.

% In Appendix~\ref{apx:}, we further  show that adopting the identical prompting method for BERT with VLMs of CLIP also shows similar results in Table~\ref{tab:visual_results} and \ref{tab:embodied_ret}.

\begin{figure}[t!]
    \centering
    \includegraphics[width=0.9\linewidth]{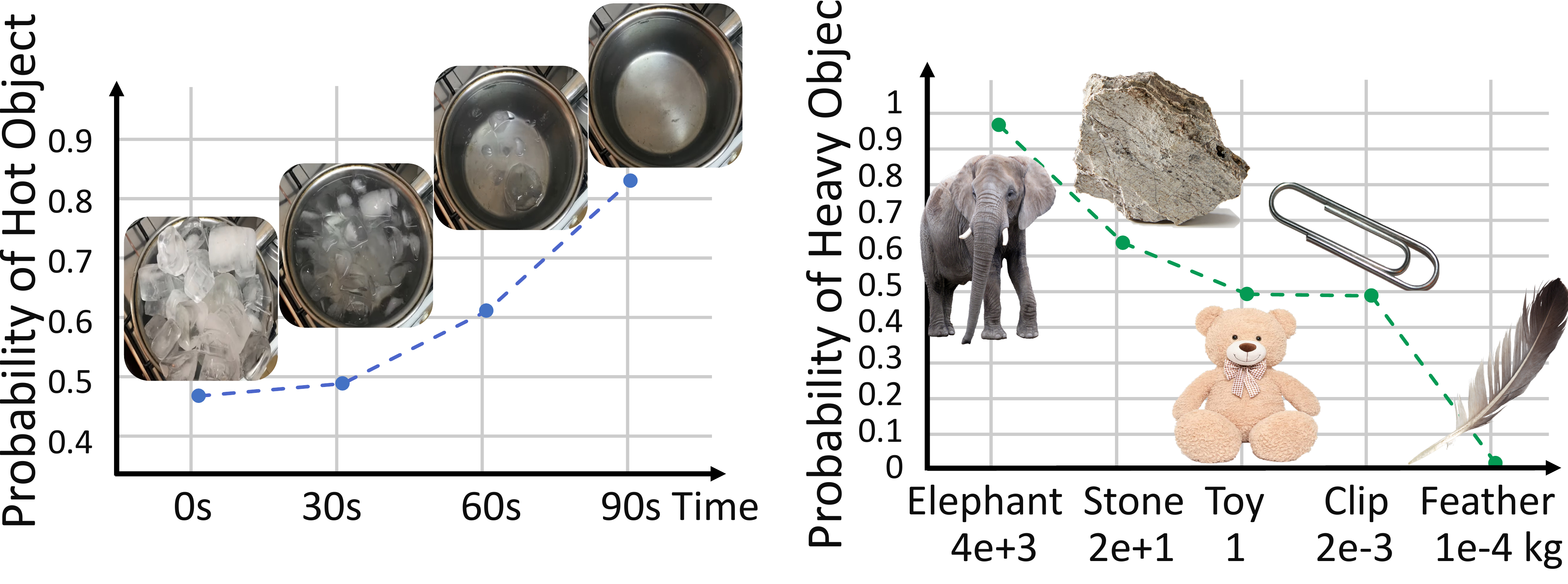}
    \caption{Case study showing that the image representations in CLIP exhibit embodied knowledge. 
    (Left) The probability of an image being classified as "hot" increases as the ice melts being heated in a boiler in a video. (Right) The probability of an image being classified as "heavy" along with corresponding mass annotation.
    }
    \label{fig:case_study}
\end{figure}

\paragraph{Learned embodied knowledge in image representations}
We are interested in how the VLMs of CLIP learn embodied knowledge. 
A potential answer is that the images contain rich semantics regarding embodied knowledge such as the heat of the object, and the knowledge can be propagated to the VLMs via the contrastive learning objective. To examine this, we perform a case study by calculating the attribute similarities among the images.
We first take clips from a video of heating a pile of ice and then perform a binary classification by calculating the cosine similarities with text prompts \emph{a photo of a hot object.} and \emph{a photo of a cold object} for each frame.
The left of Figure~\ref{fig:case_study} shows that the probability of a hot object increases during the heating procedure. Similarly, we perform a binary classification over heavy and light-weight objects ranging from an elephant to a feather. The right of Figure~\ref{fig:case_study} 
shows that the image representations are aware of the mass of different objects.
This qualitative study shows that image representations are the potential source of embodied knowledge.

\paragraph{Transferring embodied knowledge from VLMs to LMs} 
We further verify whether the learned embodied knowledge in CLIP could be transferred to text-only models.
Specifically, we perform knowledge distillation~\citep{hintonKD} by treating the original text-only language model as a student, and the CLIP text encoder as a teacher model providing the learned embodied knowledge. However, our preliminary study in Appendix~\ref{apx:other_vl_models} shows that vanilla alignment on the predicted word distributions could not be effective.
Inspired by our case study showing that the rich embodied knowledge contained in the representations,
we utilize Neuron Selectivity Transfer~\citep{huang2017nst} which transfers the inner states such as spatial
activation patterns of teacher neurons to student neurons, by aligning the token representations of the last layer between the teacher and student language models, which is implemented as a squared maximum mean discrepancy~(MMD) with a polynomial kernel to measure the distance between the activation patterns of student neurons and teacher neurons. 
The total training objective of the language model is a combination of the original language modeling loss and the MMD loss with a balancing coefficient. We refer readers to Appendix~\ref{apx:entity_distil} for more details.
As shown in Table~\ref{tab:kd_ret}, the distillation provides a performance boost on embodied concepts understanding, e.g., learning from a CLIP-ViT-L/14 teacher model achieves improvement that is comparable with that brought by scaling the model parameter $134$x from OPT-1.3B to OPT-175B.\footnote{Only OPT is adopted for experiments as the CLIP encoder cannot deal with the mask
tokens introduced in BERT.}
It validates our assumption and indicates that future studies could utilize the richer representations in VLMs for improving LMs, yet the gap between distilled LM and VLM suggests that there is still room for advancement.
% the understanding of embodied concepts.
% Nevertheless, the performance gap after distillation indicates that 
% that vision supervision helps the VLMs learn embodied knowledge in the representation,

% Besides, it can be found that better performing CLIP models.

% We note that the achievement is not marginal, as the OPT-175B onl

% we use the squared maximum mean discrepancy (MMD) [26] with kernel trick to
% measure the distance between the activation patterns of student neurons

% \begin{figure}
%      \centering
%      \begin{subfigure}[b]{0.48\linewidth}
%          \centering
%          \includegraphics[height=0.1\textheight]{iclr2023/plots/case_study_temp.pdf}
%          \caption{Heating water during time.}
%          \label{fig:case_temperature}
%      \end{subfigure}
%      \hfill
%      \begin{subfigure}[b]{0.48\linewidth}
%          \centering
%          \includegraphics[height=0.1\textheight]{iclr2023/plots/case_study_mass.pdf}
%          \caption{Probabilities of objects classified as heavy.}
%          \label{fig:case_mass}
%      \end{subfigure}
% \caption{The image representations of CLIP exhibit knowledge of temperature and mass.}
%  \label{fig:case_study}
% \end{figure}

% Possible remedies include devising better prompting methods like adaptively adjusting the prompts to eliminate ambiguity and developing better pre-training objectives taking the variations of text into consideration can also be promising.

% Perform error analysis of the CLIP model 

\paragraph{VLMs perform poorly when dealing with ambiguous text descriptions}
\label{subsec:clip_adj}

\begin{figure}[t!]
    \centering
    \includegraphics[width=0.9\linewidth]{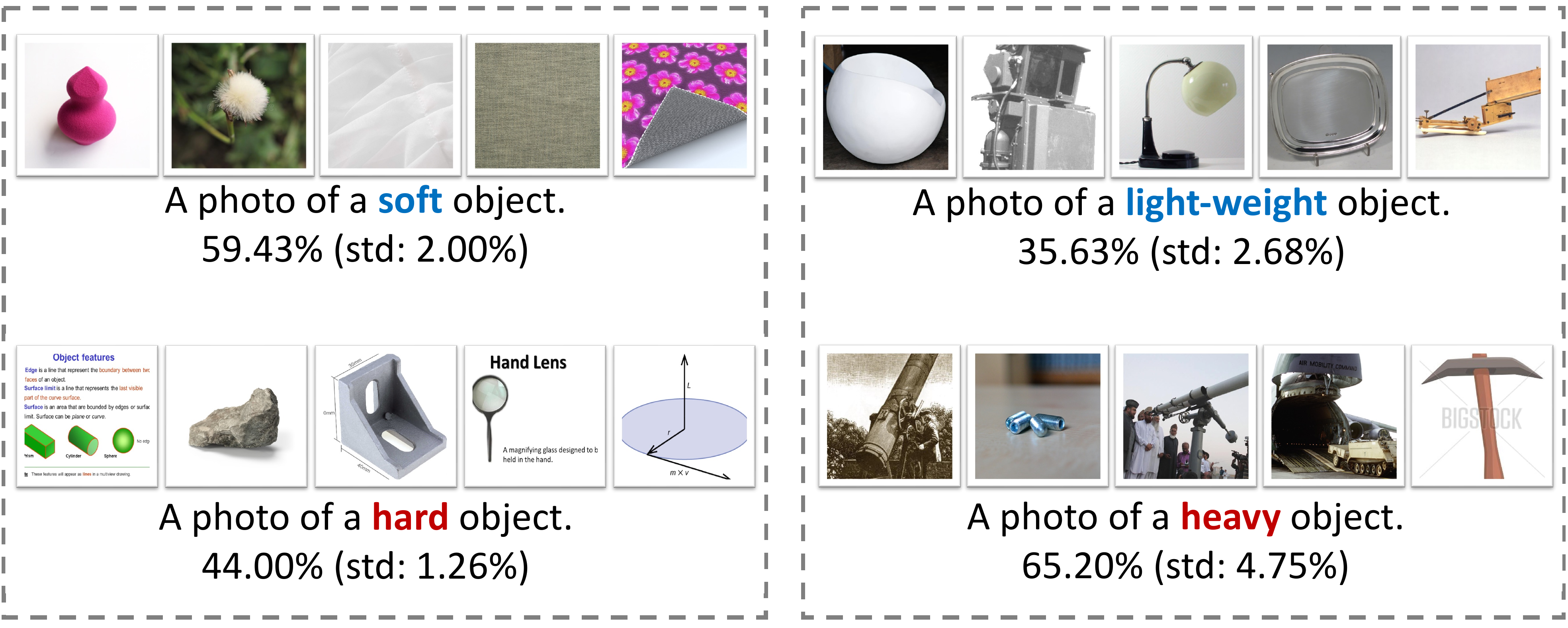}
    \caption{Top-5 retrieved images  and the prediction accuracy with different attribute prompts. 
    The accuracy drops when the text inputs contain ambiguous words and compound words, as the retrieved images are biased toward specific meanings.
    }
    \label{fig:error_analysis}
\end{figure}
% During our preliminary study, we observe that the VLMs of CLIP perform relatively worse for specific adjectives like \emph{hard}. 
% We further investigate this issue by checking the retrieved images with prompts with different attribute adjectives, on the CC12M dataset~\citep{changpinyo2021cc12m}. The results are illustrated in Figure~\ref{fig:error_analysis}.
% We find that for the text \emph{a photo of a hard object}, the retrieved images are mostly about learning materials that are abstract and difficult, with only one rock image related to the hardness.
% Besides, for the text with the compound adjective word \emph{light-weight}, the retrieved images are biased to the meanings related to lighting-bulb and light-toned color instead.
% Accordingly, the results with ambiguous texts are much lower.
% These observations show that handling the semantic ambiguity is still challenging for VLMs and explains the performance gain brought by the self-supervision in DeCLIP.

During our preliminary study, we observe that VLMs of CLIP perform relatively poorly for specific adjectives such as \emph{hard}.
To further investigate this issue, we examine the retrieved images using prompts with different attribute adjectives on the CC12M dataset~\citep{changpinyo2021cc12m}. Our results, illustrated in Figure~\ref{fig:error_analysis}, revealed that for the prompt \emph{a photo of a hard object}, the retrieved images were mostly about abstract and difficult learning materials, with only one rock image related to the attribute of hardness. 
Additionally, for the prompt \emph{light-weight}, the retrieved images are biased towards meanings related to lighting bulbs and light-toned colors. These observations demonstrate that handling semantic ambiguity remains a challenge for VLMs, suggesting that future improvements may incorporate more language-side supervision such as DeCLIP.

\section{Related Work}
\paragraph{Probing Language Models}
Understanding what LMs know after large-scale pre-training is an active research area~\citep{rogers-etal-2020-BERTology}. Various probing methods have been developed~\citep{tenney2019probing,LAMA}, and investigations show that LMs capture linguistic~\citep{tenney2019bertpipeline,liu2019linguistic}, factual~\citep{LAMA,AdamRoberts2020HowMK,dai2022knowledge}, commonsense knowledge~\citep{wang-etal-2019-make,forbes2019neuralphysical}, and even acquire grounded concepts~\citep{patel2022mapping}. 
For VLMs, studies demonstrate their potential in acquiring spatial commonsense~\citep{zhang-etal-2022-visual,liu-etal-2022-spatial-commonsense} and color perception~\citep{abdou-etal-2021-languagecolor}, yet performing worse on NLU tasks~\citep{tan2020vokenization} and achieving no significant on lexical grounding~\citep{yun2021lexicalGrounding}. 
In this paper, we investigate the understanding ability of LMs on physical concepts. 
Different from PIQA~\citep{bisk2020piqa} consists of questions requiring physical commonsense reasoning, our VEC benchmark examines the understanding ability of the most fundamental physical concepts.
The evaluation on the VEC benchmark demonstrates that text-only LMs can learn specific visual concepts after scaling up while struggling with the embodied concepts.
% We observe similar results in our LiVE benchmark when evaluating other VLMs such as Vokenization~\citep{tan2020vokenization} in Appendix~\ref{apx:other_vl_models}.

\begin{table}[t!]
    \centering
    \resizebox{\linewidth}{!}{
\begin{tabular}{@{}l|ccc|c@{}}
\toprule
\textbf{Model} & \textbf{Mass} & \textbf{Temperature} &  \textbf{Hardness} & \textbf{Avg.} \\
\midrule 
CLIP-ViT/B-32~(T1)  & 65.20\cpm{ 4.75 }   &  60.28\cpm{ 6.83  }  &  59.43\cpm{ 2.00 }  &    61.64  \\
CLIP-ViT/L-14~(T2)   &  73.15\cpm{ 6.34}  &    65.88\cpm{	2.31  }  &  69.57\cpm{	2.26} &  69.53 \\
\midrule
OPT-1.3B  & 50.05\cpm{ 0.10 }  & 50.90\cpm{  5.08 } &  53.03\cpm{ 2.69}  &  51.33   \\ 
 \quad scale up to 13B &   50.14\cpm{ 0.36 }  & 51.85\cpm{  6.34 }&  52.38\cpm{  3.09 }&  51.46 \weakimpro{0.13}\\ 
\quad scale up to 175B &   50.21\cpm{ 0.24}  &  59.83\cpm{  8.68 } &  57.33\cpm{ 3.41 } &  55.79    \impro{4.46}  \\ 
\midrule 
OPT$_\text{YFCC-15M}$ 	&50.02\cpm{ 0.05} &	57.73\cpm{ 2.24} &	50.04\cpm{ 2.98 }	& 52.61  \\ 
\quad  Distill w/ T1	& 49.88\cpm{ 0.37}  &	55.76	\cpm{ 4.01 }&	 \textbf{53.23\cpm{ 	3.12}}	& 52.96 \weakimpro{0.35}  \\
 \quad Distill w/ T2 & \textbf{54.27\cpm{ 5.20}} &	\textbf{60.78\cpm{4.23}}  &	52.91\cpm{	1.62}& \textbf{55.99}  \impro{3.38} \\ 
\bottomrule
\end{tabular}}
  \caption{Results of embodied distillation. Transferring embodied knowledge from CLIP-ViT to OPT brings a gain of 3.38 points, which is comparable with the improvements by scaling the model from 1.3B to 175B.
  % $\Delta$ denotes the performance gain by scaling model up or learning from VLMs.
  }
\label{tab:kd_ret}
\end{table}

\paragraph{Vision-Language Pre-training}
% The interest has grown recently in vision-language pre-training for unifying cross-modal representations.
Unifying cross-modal representations via vision-language pre-training has achieved promising progress.
Pilot studies adopt masked reconstruction to learn shared representations across modalities from a mixed visual and language inputs~\citep{Li2019VisualBERT,lxmert,Su2020VLBERT,Chen2019UNITER,Li2020OscarOA}. 
CLIP~\citep{Radford2021CLIP} introduces a contrastive language-image pre-training framework, utilizing language as supervision for learning transferable image representations with large-scale image-text pairs, triggering a series of variants for further improvements~\citep{Jia2021ALIGN,declip,filip,albef,li2022blip}.
Our study uses VLMs of CLIP and its variants to investigate the impact of visual supervision on understanding physical concepts and our results suggest that visual supervision is crucial for LMs to understand embodied concepts.
\section{Conclusion}
In this paper, we introduce \textbf{VEC} for evaluating the understanding of physical concepts in LMs.
Our results show that large LMs understand specific visual concepts but struggle with embodied knowledge.
VLMs instead perform much better in both the visual and the embodied world, indicating that visual signals are vital for understanding physical concepts. 
Further analysis suggests that transferring the VLM representations 
to LMs is effective for boosting embodied concepts understanding.
% , shedding light on directions for improving LMs.
% future studies.
% of  concepts.
% sheds light on future directions for further improvements on LMs.
% The further qualitative analysis demonstrates that the learned embodied knowledge is potentially from the image representations and CLIP struggles when dealing with ambiguous text inputs, shedding light on future directions for further improvements.

\section*{Limitations}
%Our study has two main limitations:
%\noindent\textbf{Narrow scope of examined embodied concepts.} Our evaluation setup in embodied knowledge is limited due to the narrow scope of types of probed knowledge.
We evaluate three physical properties, including mass, temperature, and hardness in this work. 
These aspects are selected as they could be measured by well-established metrics and easily sensed by humans. While this is a biased approximation, it is still sufficient to demonstrate that the current text-only LMs perform poorly regarding embodied knowledge and suggest that vision supervision could help understand embodied concepts. We will dynamically update our benchmark by incorporating new tasks in the future.

%\noindent\textbf{Only CLIP is adopted for the evaluation of VLMs.} 
While there exist many multi-modal models, we only adopt VLMs of CLIP and its variants for our investigation. We select CLIP due to its superior image representation performance and support text-only encoders. %, there are two reasons why it is adopted.
As our evaluation is language-oriented, the evaluated models are supposed to be able to deal with inputs with pure text. VLMs such as UNITER~\citep{Chen2019UNITER} only support multi-modal inputs and thus cannot be selected.
%(ii) The pre-training of CLIP only adopts the contrastive image-text matching objective, ablating the effect of other objectives that might dilute the contribution of visual signals.
Therefore, CLIP is chosen as a representative work of VLMs for evaluation. 
However, the findings of CLIP may not transfer to other V+L models trivially, as CLIP utilizes million-level image-text pairs collected from the web, which could also become a significant source of embodied knowledge.
Besides, there are recently proposed VLMs models with various architectures and pre-training recipes such as SimVLM~\citep{wang2021simvlm}, UniT~\citep{hu2021unit}, ViLT~\citep{kim2021vilt} and FLAVA~\citep{singh2022flava}, which show promising performance in both cross-modal and single-modality tasks. We will examine these advanced models in our benchmark in the future. 

\bibliography{custom}
\bibliographystyle{acl_natbib}
% \clearpage

\appendix

% Add few-shot fine-tuning on W3
\section{Matching-based Prompting Results for BERT}
\label{apx:bert_cls}

We perform the identical prompting method and templates  with CLIP VLMs for LMs of BERT. RoBERTa is discarded as there is no pooled embedding  during its pre-training for representing the sentence.
As shown in Table~\ref{tab:bert_pooled_output}, the performance is still close to a random guessing baseline. Besides, the higher variances on all the tasks compared to VLMs of CLIP indicate that this method does not fit LMs of BERT well, validating our prompting design choice in the main paper to fit the pre-training paradigm.

\section{Model Configurations}
\label{apx: model_parameters}

\begin{table*}[t!]

\centering

\scalebox{0.85}{
\begin{tabular}{@{}l|cccc@{}}
\toprule
\textbf{Model} &  \textbf{Hidden Layers} &  \textbf{Hidden Size } &  \textbf{Attention Heads} & \textbf{Total \# of Parameters} \\

\midrule
BERT-base      &  12 & 768 & 12  &   110M \\
BERT-large     &  24 & 1,024 & 16  &  340M \\
RoBERTa-base   &  12 & 768 & 12  &  125M \\
RoBERTa-large    &  24 & 1,024 & 16  & 355M \\
%	%
\midrule 
OPT-125M   &  12 & 768 & 12 & 125M\\ 
OPT-1.3B  & 24 & 2,048 & 32&  1.3B\\ 
OPT-13B   & 40 & 5,120 & 40 & 13B\\ 
OPT-175B   & 96 & 12,288 & 96&  175B\\ 
\midrule 
CLIP-ViT/B-32  & 12 & 512 & 8  & 63M\\
% CLIP-ViT/B-16  &      &   &       &         & &  \\
DeCLIP-ViT/B-32  & 12 & 512 & 8 & 63M\\
CLIP-ViT/L-14& 12 & 768 & 12 &123M  \\

\bottomrule
\end{tabular}}
\caption{Detailed configuration of models evaluated in the paper. }
\label{tab:model_parameters}
\end{table*}

Here we provide the detailed configurations of evaluated LMs and VLMs in our main paper.
For the vanilla model, there model configurations are listed in Table~\ref{tab:model_parameters}.
For the models pre-trained from scratch with the YFCC-15M dataset, they are adopted the same LM architecture as the VLM of CLIP-ViT/B-32, and the only difference is the pre-training objective, as shown in Table~\ref{tab:w1_model_config}.
All these models are optimized using an Adam optimizer with a learning rate set to $1$e$-4$, linearly increased at the first $2000$ steps. The batch size is $2048$ and all models are trained with $32$ epochs. We use 1\% of the data for evaluation, and the final OPT-YFCC15M model gets a $31.9$ validation perplexity.
% We provide the detailed configurations of models we evaluated in our main paper.
% Specifically, for the evaluation in \Wone, we train identical Transformer~\citep{vaswani2017attention} language models configured with $12$ Transformer layers and $512$ hidden units with $8$ attention heads, from scratch on the YFCC15M dataset.
% These models only differ in the pre-training objectives, as shown in Table~\ref{tab:w1_model_config}.

\begin{table*}[t!]
\centering

\scalebox{0.85}{
\begin{tabular}{@{}l|cc@{}}
\toprule
\textbf{Model} &  \textbf{Training Objective}  &  \textbf{Training Dataset}\\

\midrule
BERT$_\text{YFCC-15M}$ &   Masked Language Modeling~(MLM) & Captions in YFCC-15M \\
GPT$_\text{YFCC-15M}$ &   Causal Language Modeling~(MLM) & Captions in YFCC-15M \\
CLIP$_\text{YFCC-15M}$   & Contrastive Image-text Matching~(CIM) & Image-Text Pair in YFCC-15M\\
% DeCLIP &  MLM + CIM &  Image-Text Pair in YFCC-15M\\
% DeFILIP &  MLM + Contrastive Token-Patch Matching & Image-Text Pair in YFCC-15M\\
\bottomrule

\end{tabular}}
\caption{Pre-training objectives and corpus comparison of YFCC-15M models evaluated in the main paper.}
\label{tab:w1_model_config}
\end{table*}

\section{Details of Prompts}
\label{apx:prompts}

We provide the used prompts for evaluating different models based on their pre-training objectives. Examples of Head, Rel and Tail of each dataset are shown in Table~\ref{tab:all_dataset_statistics}. Due to the sensitivity of language models to prompts, we provide diverse prompts for each model on each task.

\paragraph{Prompts for Masked Language Models} A \texttt{[MASK]} token is placed in the prompt and the models are asked to predict the probabilities of the \texttt{[MASK]} token. To avoid multiple mask tokens in prompts, we follow \citeauthor{timoPET} to convert knowledge fact into a cloze-question. For example,  a temperature fact \texttt{(water, colder than, frying oil)} can be converted into \texttt{Q: is the water colder than frying oil? A: [MASK]!}. The models need to choose the token \texttt{yes} or \texttt{no} to fill the mask.

\paragraph{Prompts for Causal Language Models} As there is no special \texttt{[MASK]} token during the pre-training of causal language models, we do not use \texttt{[MASK]} tokens in prompts for causal language models. For Color, Shape and Material datasets of the visual concepts we construct two prompts for (Head, Tail$_1$) and (Head, Tail$_2$); while for other datasets, we construct two prompts for (Head, Rel, Tail) and (Head, Rel$'$, Tail) where Rel$'$ is the antonym relation of Rel. The prediction is based on the prompt with lower perplexity.

\paragraph{Prompts for CLIP} Following \citet{Radford2021CLIP}, we use prompts like \texttt{a photo of ...} here. As the language encoder of CLIP encodes sentences to a vector and can evaluate similarities between sentences. We use an attribute prompt like \texttt{a photo of a cold object} and construct same prompts for objects (water and frying oil) in the knowledge fact. We can determine the colder object if the prompt of this object has a higher similarity to the attribute prompt.

\begin{table*}[hbt!]
\centering

\scalebox{0.8}{
\begin{tabular}{@{}l|cccccccc|c@{}}
\toprule
\textbf{Model} & \multicolumn{2}{c}{\textbf{SST-2}}   & \multicolumn{2}{c}{\textbf{QQP}}  &  \multicolumn{2}{c}{\textbf{QNLI}} & \multicolumn{2}{c|}{\textbf{MNLI (m / mm)}} & \textbf{Avg.} \\ 
\midrule 

BERT~(Wiki) & \multicolumn{2}{c}{90.13}  & \multicolumn{2}{c}{  83.20}  & \multicolumn{2}{c}{87.57}   & \multicolumn{2}{c|}{ 78.90 / 80.05}  & 83.97 \\ 
DistilledOscar &  \multicolumn{2}{c}{89.33}   &  \multicolumn{2}{c}{67.98}    &   \multicolumn{2}{c}{82.48}   &   \multicolumn{2}{c|}{  74.46 / 74.82} &  77.81  \\  
VLM-BERT-base  & \multicolumn{2}{c}{90.60}  &  \multicolumn{2}{c}{ 90.10} & \multicolumn{2}{c}{89.47}&  \multicolumn{2}{c|}{  81.57 / 82.43} & 86.83 \\ 
VLM-RoBERTa-base & \multicolumn{2}{c}{90.13}  &  \multicolumn{2}{c}{88.44}             & \multicolumn{2}{c}{87.91}   & \multicolumn{2}{c|}{ 80.37 /  80.43}&  85.46\\ 
  \midrule 
 \textbf{Model} & \textbf{Color} & \textbf{Shape} & \textbf{Size} & \textbf{Height} & \multicolumn{1}{c|}{\textbf{Material}} &\textbf{Mass} & \textbf{Temperature} &  \textbf{Hardness} & \textbf{Avg.} \\
\midrule
BERT~(Wiki) & 49.41 &48.07  & 51.70 &  49.46&\multicolumn{1}{c|}{52.39} & 48.85 & 51.07 & 52.34 &  50.41 \\ 
DistilledOscar &  49.97   &  53.61	       &  49.07   &  49.80
  & 	 \multicolumn{1}{c|}{51.46}  & {51.22}   &  47.94  &  51.23  &  50.54  \\ 
VLM-BERT-base  & 50.69   &   50.07       & 51.00  &  50.92
  &  \multicolumn{1}{c|}{53.89} & 44.83  & 50.64  & 49.22  &  50.16 \\ 
VLM-RoBERTa-base &  49.53 &  51.21              & 49.00   & 
 49.22  &  \multicolumn{1}{c|}{49.54}  & 49.92  &  51.11   &  49.63  &  49.90 \\ 
% \midrule 
% BERT-base Pooled  &   &              &     & 
%   &  \multicolumn{1}{c|}{}  &   &     &   &    \\ 
%   BERT-large Pooled &   &              &     & 
%   &  \multicolumn{1}{c|}{}  &   &     &   &    \\ 
\bottomrule
\end{tabular}}

\caption{Fine-tuned accuracy of other visual-informed pre-trained language models on NLU tasks and zero-shot results regarding the physical concepts.}
\label{tab:other_visual_model}
\end{table*}

\begin{table*}[hbt!]
\centering

\resizebox{\linewidth}{!}{
\begin{tabular}{@{}l|cccccccc|c@{}}
\toprule
 \textbf{Model} & \textbf{Color} & \textbf{Shape} & \textbf{Size} & \textbf{Height} & \multicolumn{1}{c|}{\textbf{Material}} &\textbf{Mass} & \textbf{Temperature} &  \textbf{Hardness} & \textbf{Avg.} \\
\midrule
%	%
%	 %
CLIP-ViT/L-14 &  80.33\cpm{  3.61 } &   \textbf{85.00\cpm{   4.03}     }       &  63.96\cpm{  6.10 }  & 
 60.72\cpm{ 5.56} &   80.33\cpm{ 3.61  }  &  \multicolumn{1}{|c}{  73.15\cpm{ 6.34  } } &     65.88\cpm{	2.31	}   &   69.57\cpm{	2.26}  &  72.37  \\  
BERT-base Pooled  & 43.19\cpm{ 5.13 } & 59.64\cpm{  7.24  }           & 66.10\cpm{	7.91 }   &  65.48\cpm{ 	6.63 }
  & 46.55\cpm{ 	6.18} &  \multicolumn{1}{|c}{ 52.32\cpm{	7.46 } } &  56.59\cpm{	5.65  }&   55.51\cpm{ 5.50   }   &   55.67 \\ 
  
  BERT-large Pooled &  44.74\cpm{ 5.93 } &     56.93\cpm{  6.45 }    &  53.80\cpm{	5.92 }  & 54.84\cpm{ 	8.01 }
  &  52.18\cpm{  4.88} & \multicolumn{1}{|c}{57.92\cpm{ 	7.26 } } &  51.90\cpm{ 	6.56 } &    56.22\cpm{ 	4.45}     & 53.57   \\ 
\bottomrule
\end{tabular}}

\caption{Zero-shot results of BERT models with pooled output as sentence embedding on VEC benchmark.}
\label{tab:bert_pooled_output}
\end{table*}

% \begin{figure}[thb!]
%     \centering
% \includegraphics[width=0.9\linewidth]{iclr2023/plots/bar_gravity.pdf}
%     \caption{The center of gravity of different models in all linguistic probing tasks. The gravity is measured with the MDL compression. The linguistic information is centered in lower layers of CLIP models than BERT.}
%     \label{fig:gravity}
% \end{figure}

% \begin{figure}
%     \centering
% \includegraphics[width=0.9\linewidth]{iclr2023/plots/Layer_Alignment.pdf}
%     \caption{Layer-wise alignment score of text and visual features on CIFAR100. Lower alignment scores indicate better alignments between modalities.}
%     \label{fig:layer_alignment}
% \end{figure}

\section{Entity Analysis}
\label{apx:entity_dist}
\begin{figure}[t!]
    \centering
    \includegraphics[width=0.8\linewidth]{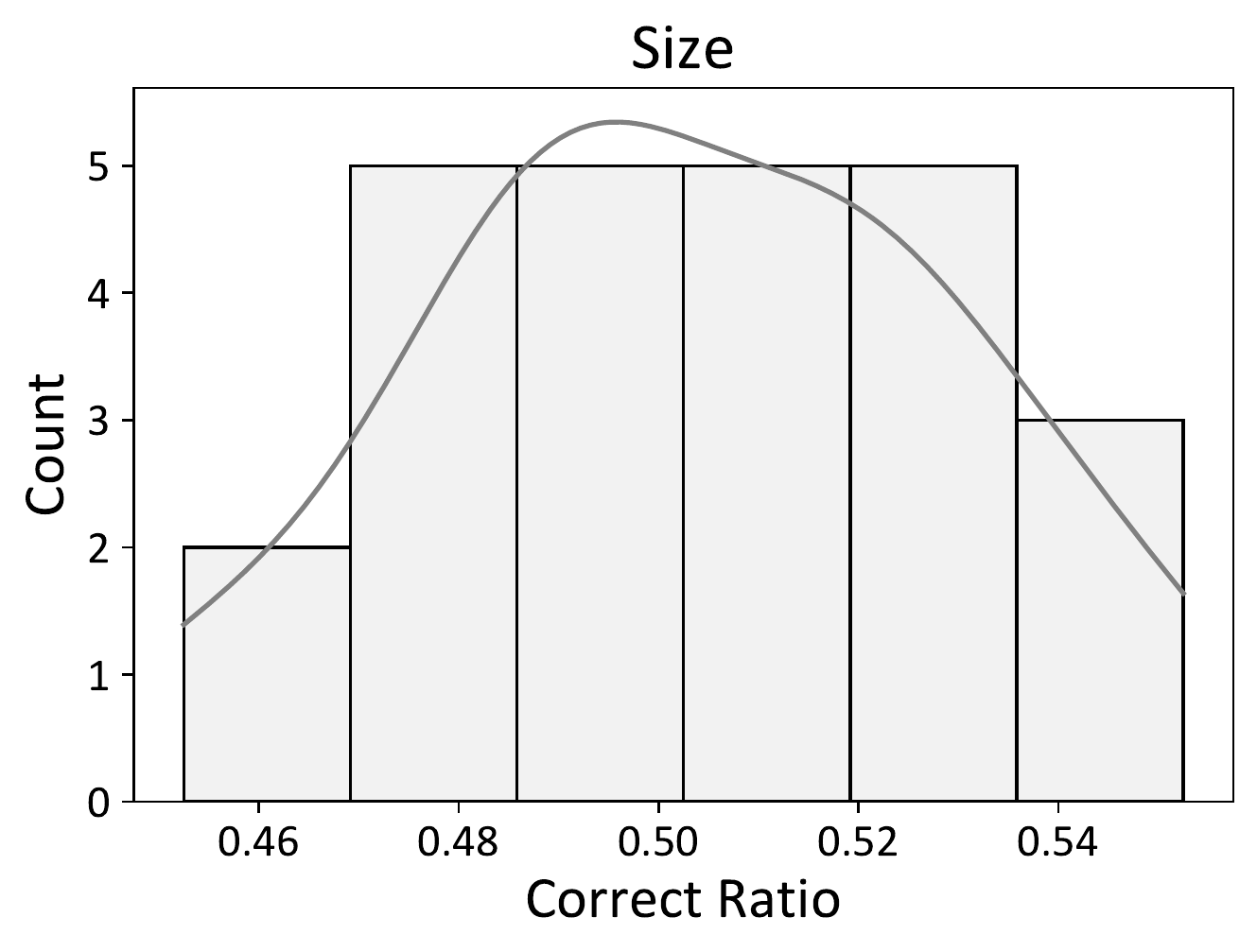}
    \caption{Histogram of entity correct ratio across different prompts on the Size dataset.}
    \label{fig:size_hist}
\end{figure}

\begin{figure}[t!]
    \centering
    \includegraphics[width=0.8\linewidth]{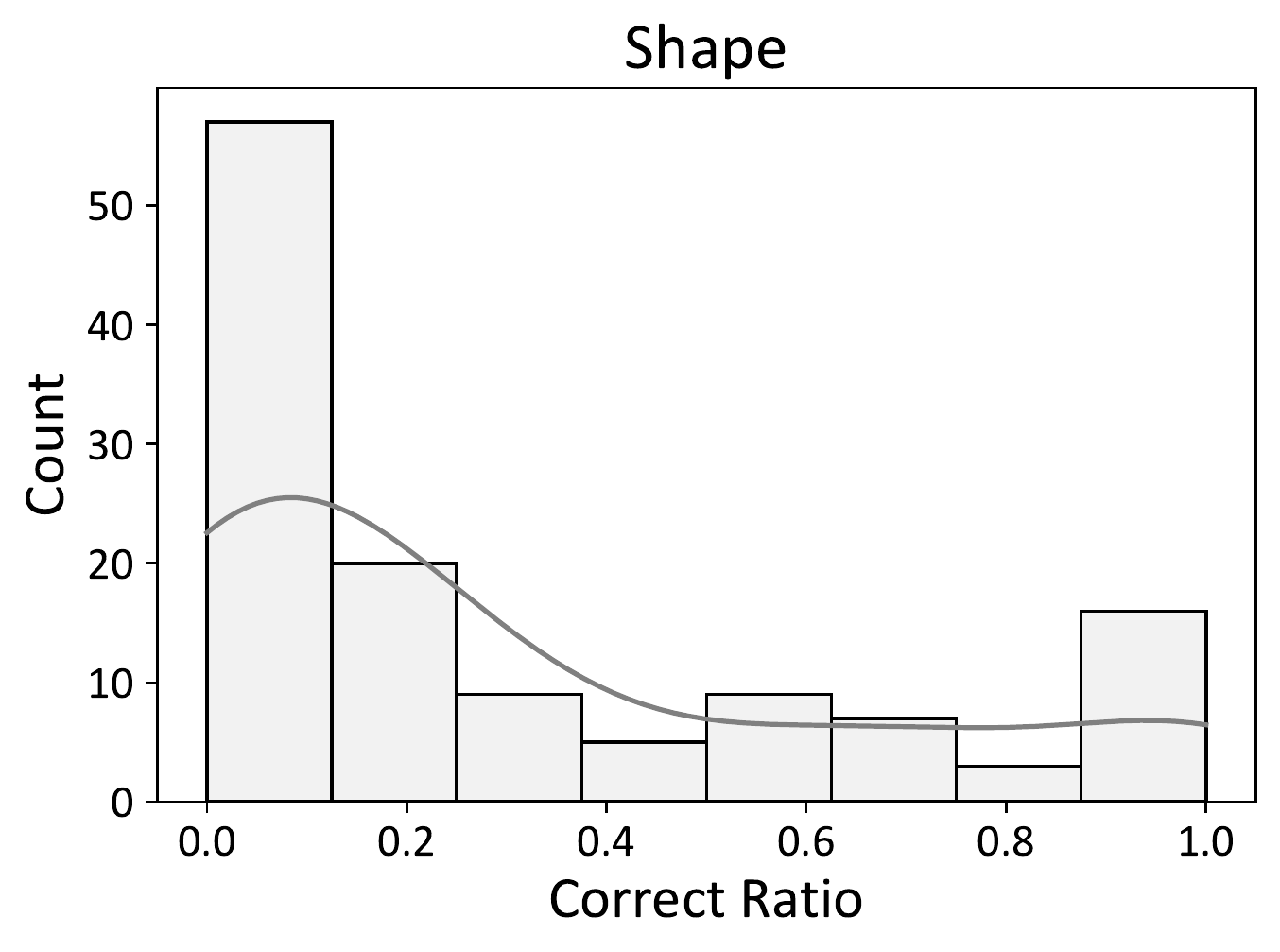}
    \caption{Histogram of entity correct ratio across different prompts on the Shape dataset.}
    \label{fig:shape_hist}
\end{figure}

\begin{figure}[t!]
    \centering
    \includegraphics[width=0.8\linewidth]{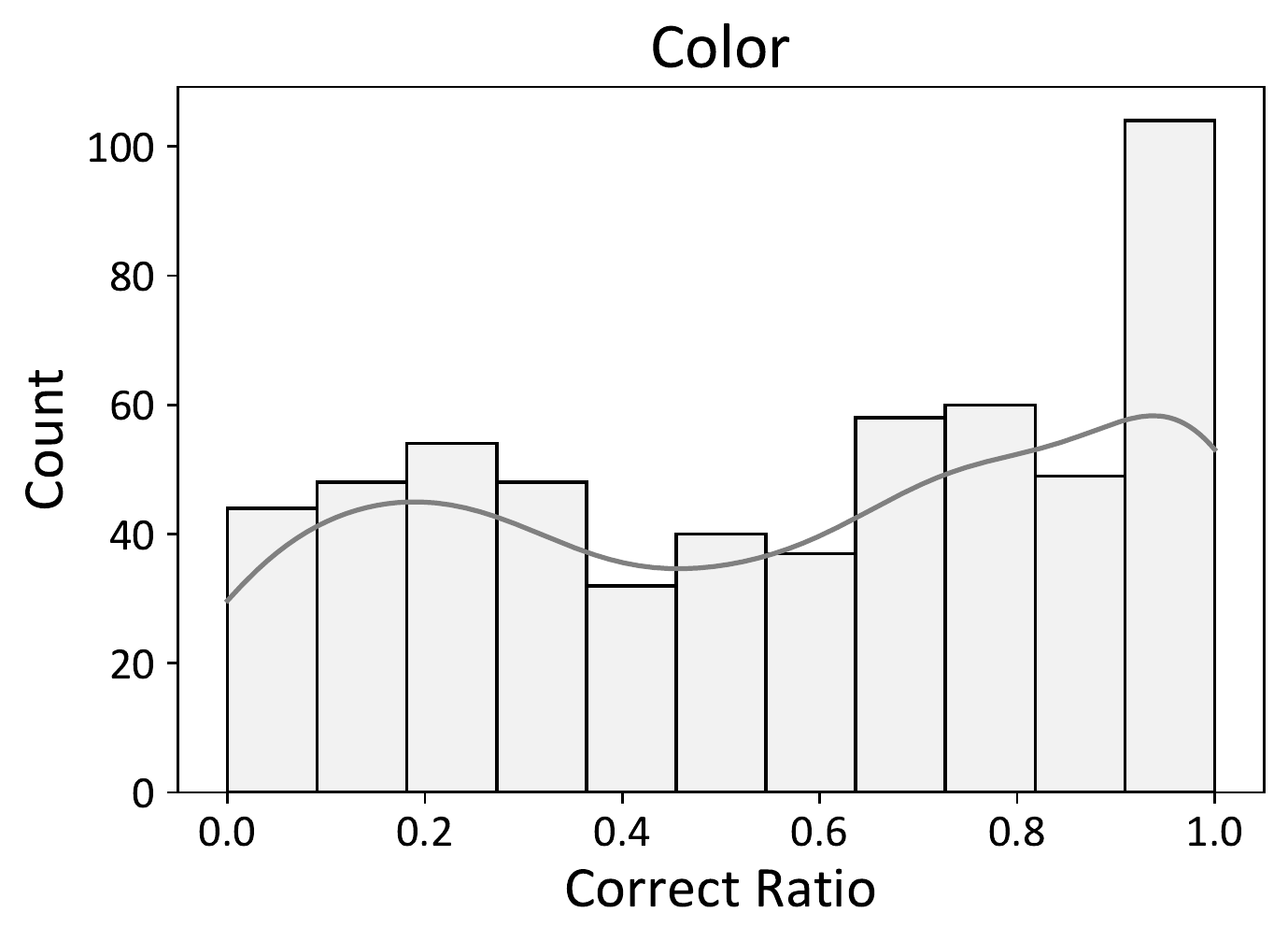}
    \caption{Histogram of entity correct ratio across different prompts on the Color dataset.}
    \label{fig:color_hist}
\end{figure}

\begin{figure}[t!]
    \centering
    \includegraphics[width=0.8\linewidth]{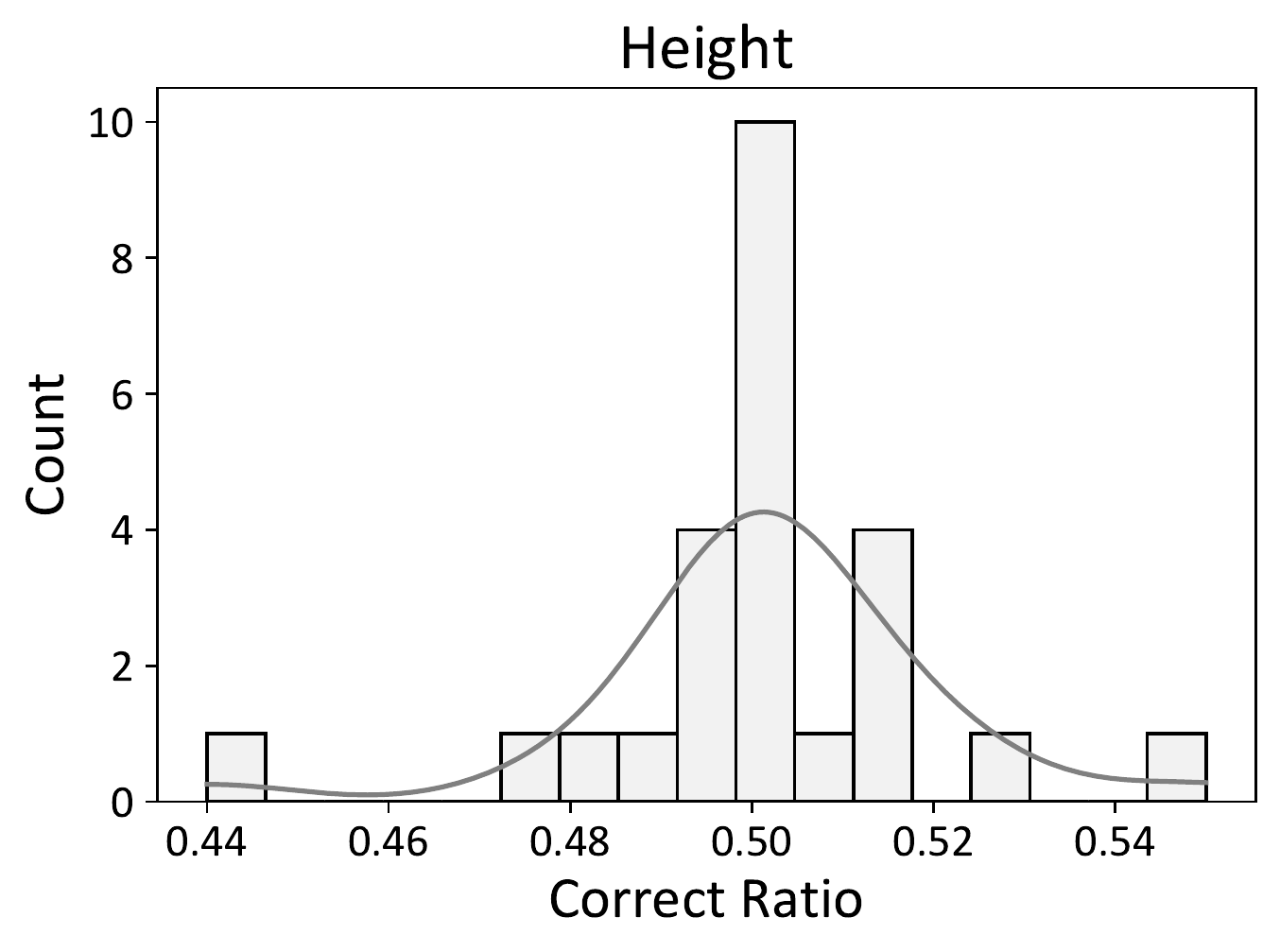 }
    \caption{Histogram of entity correct ratio across different prompts on the Height dataset.}
    \label{fig:height_hist}
\end{figure}

\begin{figure}[t!]
    \centering
    \includegraphics[width=0.8\linewidth]{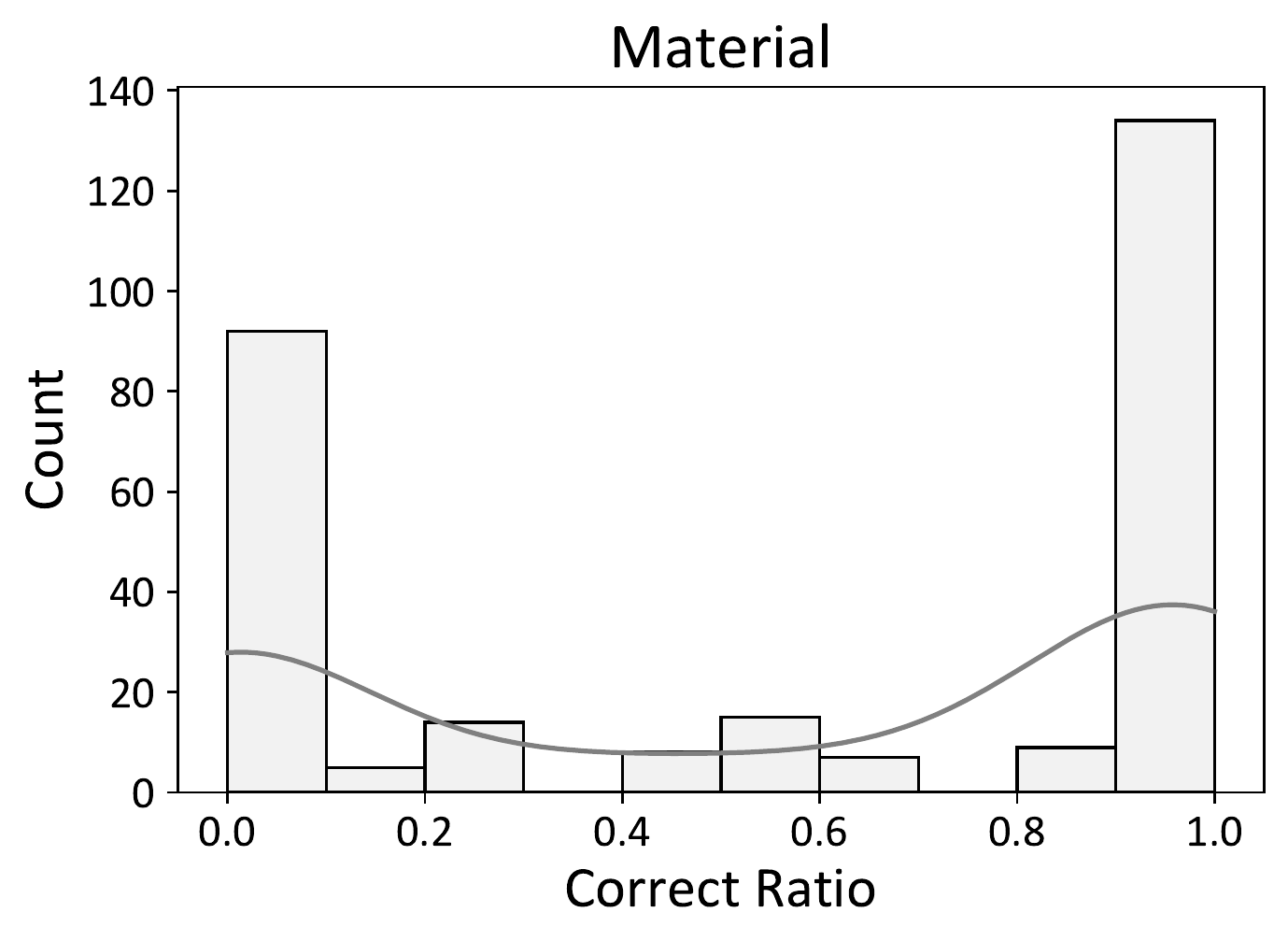 }
    \caption{Histogram of entity correct ratio across different prompts on the Material dataset.}
    \label{fig:material_hist}
\end{figure}

\begin{figure}[t!]
    \centering
    \includegraphics[width=0.8\linewidth]{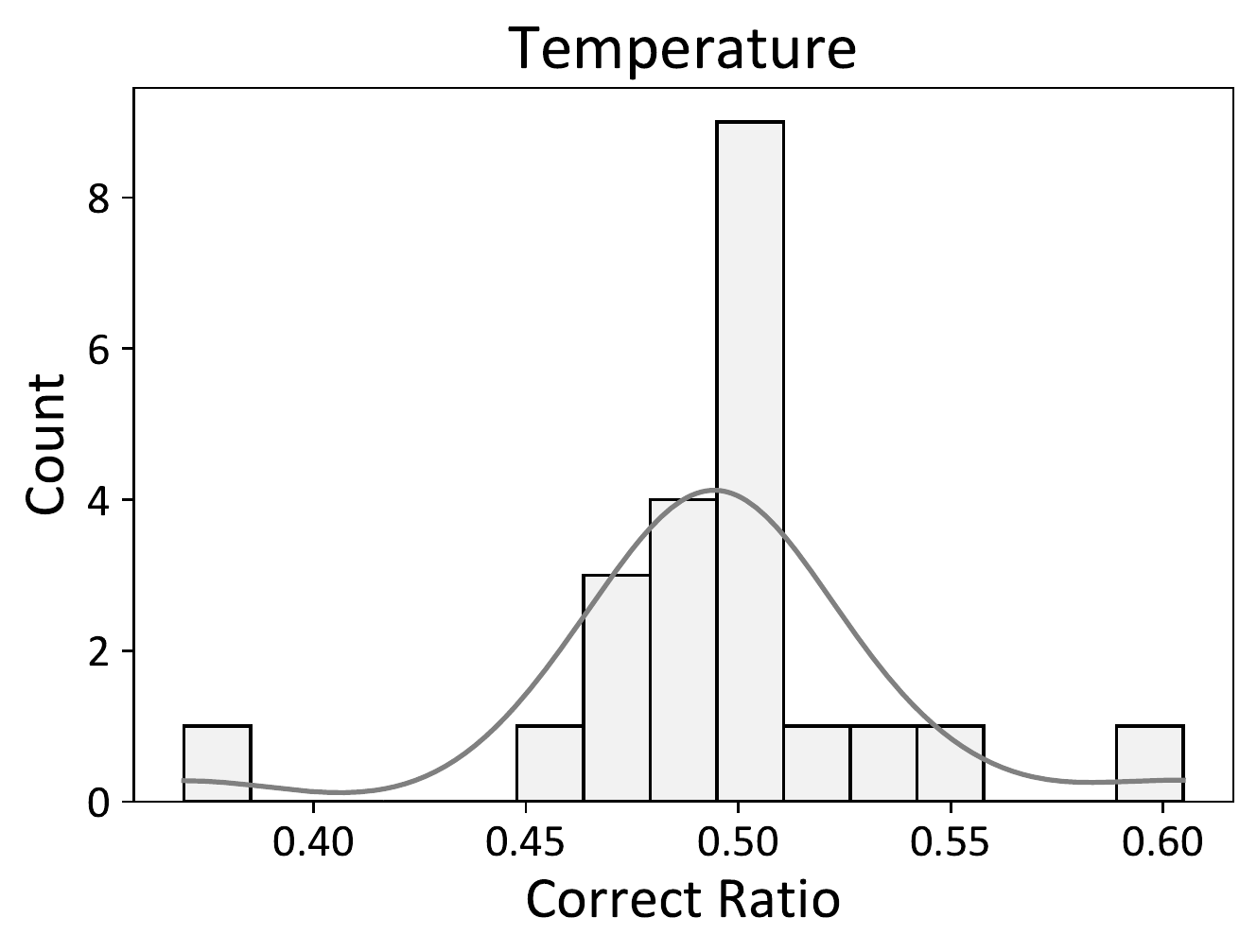 }
    \caption{Histogram of entity correct ratio across different prompts on the Temperature dataset.}
    \label{fig:temp_hist}
\end{figure}

\begin{figure}[t!]
    \centering
    \includegraphics[width=0.8\linewidth]{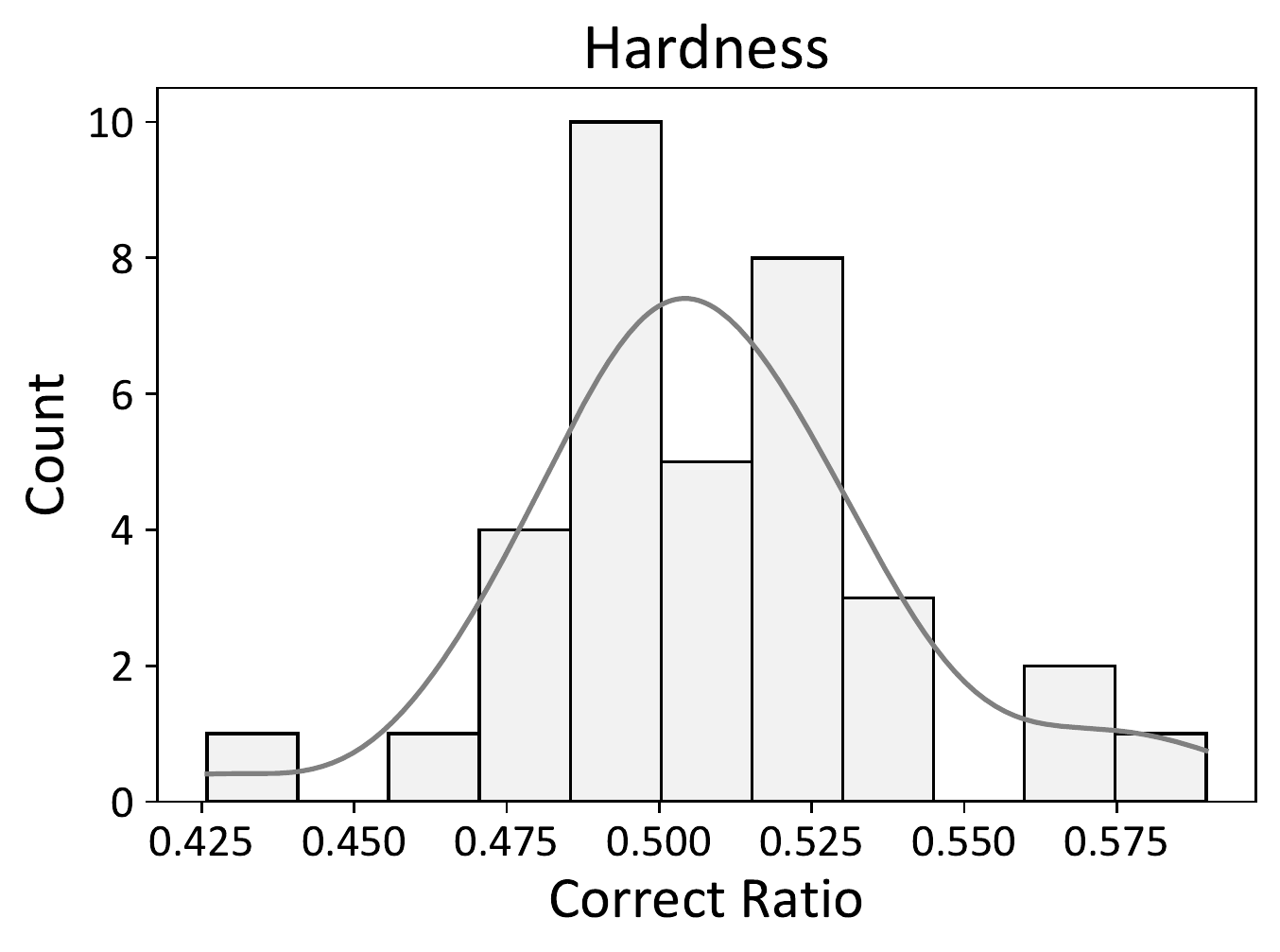 }
    \caption{Histogram of entity correct ratio across different prompts on the Hardness dataset.}
    \label{fig:hardness_hist}
\end{figure}

\begin{figure}[t!]
    \centering
    \includegraphics[width=0.8\linewidth]{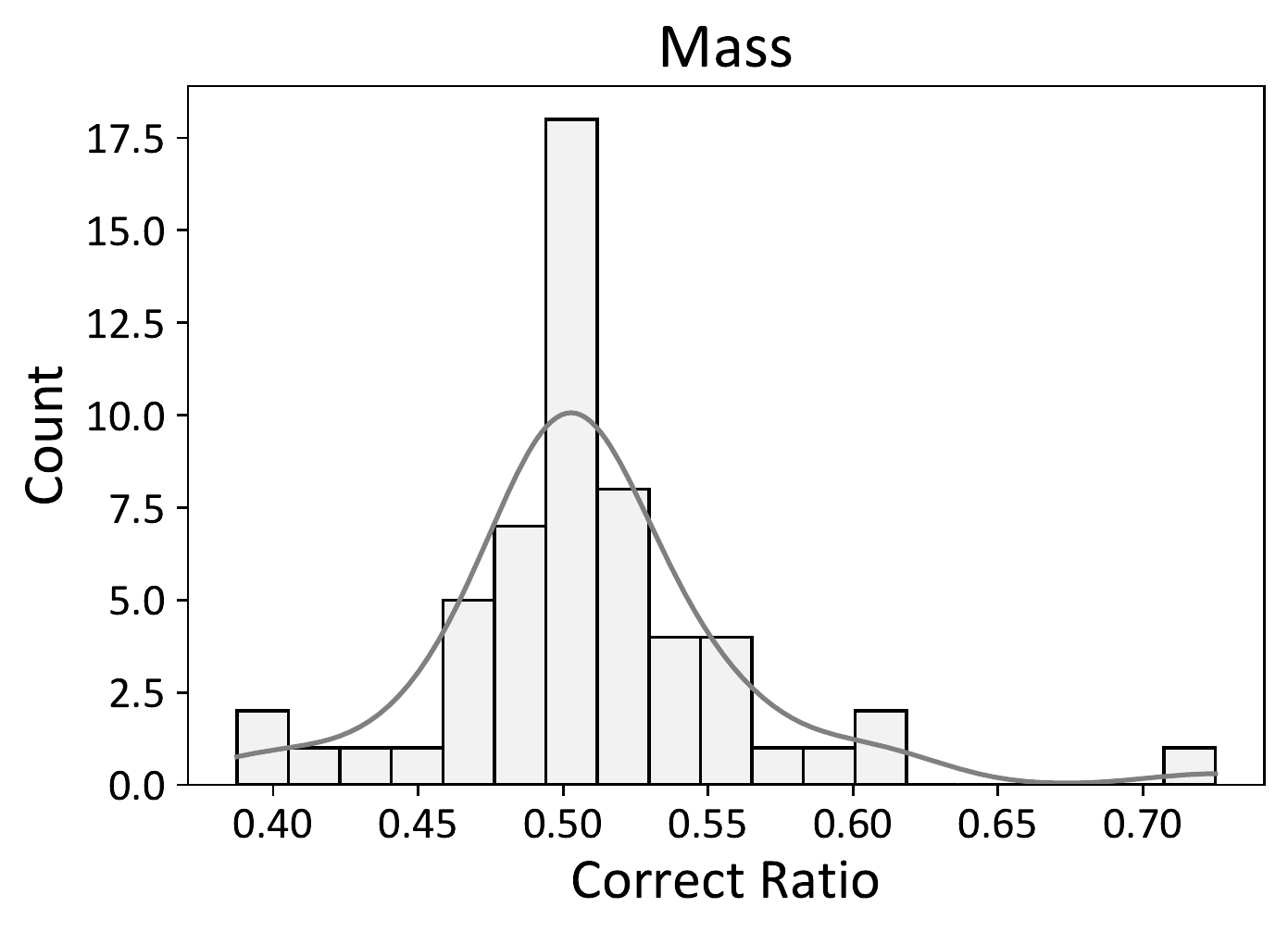 }
    \caption{Histogram of entity correct ratio across different prompts on the Mass dataset.}
    \label{fig:mass_hist}
\end{figure}

In our main paper, we investigate the random-level performance of BERT models by exploring the correct ratio over different prompts. We provide full histogram plots of all tasks in Figure~ \ref{fig:size_hist}, \ref{fig:shape_hist}, \ref{fig:color_hist},\ref{fig:height_hist},  \ref{fig:material_hist}, \ref{fig:temp_hist}, \ref{fig:hardness_hist}, and \ref{fig:mass_hist}. 
It can be found that for visual concepts tasks such as material and shape, there are entities that BERT could produce consistent correct prediction across different prompts. However, for all embodied tasks, the histograms exhibit bell curves, indicating the poor understanding ability of BERT on embodied concepts.

\section{Embodied Knowledge Transfer}
\label{apx:entity_distil}
We provide implementation details here for the knowledge transfer experiments from VLMs to LMs.
Specifically, we take the VLM as a teacher model $T$ (e.g., the text encoder of the CLIP model) and the LM as a student model $S$ (e.g., the OPT language model). 
Given a text $\mathbf{x}$ from the training dataset $\mathcal{D}$, we transfer the sequential activation patterns of $T(\mathbf{x}) \in R^{|x| \times d}$ 
to $S(\mathbf{x})\in R^{|x| \times d}$
, where $T(\mathbf{x})$ and $S(\mathbf{x})$ denote the last hidden representations of the VLM and the LM, respectively. $d$ is the number of hidden units.
The squared maximum mean discrepancy~(MMD) with kernel trick~\citep{huang2017nst} is adopted to
measure the distance between the activation patterns:
\begin{align*}
\operatorname{MMD}^2(\mathbf{x})= & \frac{1}{d^2} \sum_{i=1}^d \sum_{i^{\prime}=1}^d k\left[S(\mathbf{x})_{*, i} ; S(\mathbf{x})_{*, i^{\prime}}\right] \\ 
& +\frac{1}{d^2} \sum_{j=1}^d \sum_{j^{\prime}=1}^d k\left[T(\mathbf{x})_{*, j} ; T(\mathbf{x})_{*, j^{\prime}}\right] \\
& -\frac{2}{d^2} \sum_{i=1}^d \sum_{j=1}^d k\left[S(\mathbf{x})_{*, i} ; T(\mathbf{x})_{*, j}\right]
\end{align*}

We adopt a polynomial kernel $k(\mathbf{x}; \mathbf{y})=\left(\mathbf{x}^{\top} \mathbf{y}+c\right)^p$ with $p=2$ and $c=0$.
The MMD objective $\mathcal{L}_{\text{MMD}}$ is minimized along with the original language modeling objective $\mathcal{L}_{\text{LM}}$:
\begin{equation*}
    \mathcal{L} = \mathcal{L}_{\text{lm}} + \beta \mathcal{L}_{\text{MMD}}
\end{equation*}
where $\beta$ is a weighting factor set to $20$ to achieve a balance between objectives.
\section{Evaluation and Distillation with Oscar}
\label{apx:other_vl_models}

We examine whether other vanilla distillation from traditional V+L pre-training methods brings gains regarding visual and embodied knowledge.
Specifically, following \citet{zhang-etal-2022-visual}, we distill the knowledge of Oscar~\citep{Li2020OscarOA} into a BERT model by performing knowledge distillation~\citep{hintonKD} on the image-caption pair dataset.
Specifically, the paired text and image are fed into the Oscar model for getting the vision-aware vocabulary distribution, and a student BERT model is performing masked language modeling on the text data only and learns from the soft labels provided by the Oscar teacher model.
The distillation results in a DistilledOscar model supporting text-only inputs.
We also evaluate VLM-BERT learned via Vokenziation ~\citep{tan2020vokenization}, which devises a fine-grained token-voken matching framework to utilize visual supervision.
The models are evaluated on the four largest datasets in GLUE, including SST-2~\citep{socher2013sst}, QQP~\citep{iyer2017qqp}, QNLI~\citep{rajpurkar2016squad} and MNLI~\citep{williams2018mnli} for stable results.
As shown in Table~\ref{tab:other_visual_model}, DistilledOscar performs worse than the vanilla BERT in both NLU tasks and probing tasks regarding visual and embodied knowledge.
Besides, while VLM-BERT achieves improvements on NLU tasks, it still performs at the random level on the probed tasks.
These indicate that not all VLMs could learn embodied knowledge and it is non-trivial to distill the visual supervision from VLMs to LMs via purely language modeling.
% We think the reason is not the differences in the training objectives and the model architecture, but the data scale used for pre-training, i.e., .
% CLIP builds a $400$M paired image-text dataset for training, yet the amount for Oscar and Vokenziation is less than $10$M. 
% As recent studies suggest that purely text language models acquire complicated reasoning abilities during scaling up in model parameters and training corpus~\citep{brown2020language,wei2022emergent}, investigating the emergent abilities during the scaling up of multi-modal models.
% can also be interesting.

% Extra prompts can be used 
% [Head] is commonly found in [Tail]-shaped objects.
% What shape is typically associated with [Head]? [Tail].
% [Head] is often seen in the form of [Tail].
% The shape of [Head] is usually [Tail].
% [Head] is frequently found in objects with a [Tail] shape.
% [Head] is commonly seen in [Tail]-shaped structures.
% The typical shape of [Head] is [Tail].
% [Head] is typically associated with objects that have a [Tail] shape.
% The shape of [Head] is often [Tail].
% [Head] is commonly found in objects that are shaped like [Tail].

\begin{table*}[tbh!]
\centering
\scriptsize 
\caption{Prompts for Masked Language Models}
\begin{tabularx}{\textwidth}{c|c|l}
\toprule
Model                            & Task                                        & Prompt                                                                                                                                                                                                                                                                                                                                                                                                                                                                                                                                                                                                                                                                                                                    \\
\midrule
\multirow{40}{*}{BERT \& RoBERTa} & Size, Height, Temperature, Weight, Hardness & \begin{tabular}[c]{@{}l@{}}is the {[}Head{]} {[}Rel{]} than the {[}Tail{]}? {[}MASK{]}!\\ is the {[}Head{]} {[}Rel{]} than the {[}Tail{]}? {[}MASK{]}.\\ is {[}Head{]} {[}Rel{]} than {[}Tail{]}? {[}MASK{]}!\\ is {[}Head{]} {[}Rel{]} than {[}Tail{]}? {[}MASK{]}.\\ is {[}Head{]} {[}Rel{]} compared with {[}Tail{]}? {[}MASK{]}.\\ is {[}Head{]} {[}Rel{]} compared with {[}Tail{]}? {[}MASK{]}!\\ compared with {[}Tail{]}, is {[}Head{]} {[}Rel{]}? {[}MASK{]}.\\ compared with {[}Tail{]}, is {[}Head{]} {[}Rel{]}? {[}MASK{]}!\\ is {[}Head{]} usually {[}Rel{]} than {[}Tail{]}? {[}MASK{]}.\\ is {[}Head{]} usually {[}Rel{]} than {[}Tail{]}? {[}MASK{]}!\end{tabular} \\
\cmidrule{2-3}
                                 & Color                                       & \begin{tabular}[c]{@{}l@{}}can {[}Head{]} be of color {[}Tail{]}? {[}MASK{]}!\\ can {[}Head{]} be of color {[}Tail{]}? {[}MASK{]}.\\ is the color of a {[}Head{]} {[}Tail{]}? {[}MASK{]}!\\ is the color of a {[}Head{]} {[}Tail{]}? {[}MASK{]}.\\ is {[}Head{]} {[}Tail{]}? {[}MASK{]}.\\ is {[}Head{]} {[}Tail{]}? {[}MASK{]}!\\ is {[}Head{]} typically in {[}Tail{]}? {[}MASK{]}.\\ is {[}Head{]} typically in {[}Tail{]}? {[}MASK{]}!\\ Q: is {[}Head{]} of color {[}Tail{]}? A: {[}MASK{]}. \\ Question: is {[}Head{]} of color {[}Tail{]}? Answer: {[}MASK{]}.\end{tabular}                                                                                                \\
\cmidrule{2-3}
                                 
                                 & Shape                                       & \begin{tabular}[c]{@{}l@{}}can {[}Head{]} be the shape of {[}Tail{]}? {[}MASK{]}.\\ can {[}Head{]} be the shape of {[}Tail{]}? {[}MASK{]}!\\ does the {[}Head{]} have a shape of {[}Tail{]}? {[}MASK{]}.\\ does the {[}Head{]} have a shape of {[}Tail{]}? {[}MASK{]}!\\ is {[}Head{]} of {[}Tail{]}? {[}MASK{]}.\\ is {[}Head{]} of {[}Tail{]}? {[}MASK{]}!\\ Q: is {[}Head{]} of {[}Tail{]}? A: {[}MASK{]}.\\ Question: is {[}Head{]} of {[}Tail{]}? Answer: {[}MASK{]}.\\ {[}Tail{]} {[}Head{]}? {[}MASK{]}.\\ is {[}Head{]} typically {[}Tail{]}? {[}MASK{]}.\end{tabular}                                                                                                    \\
\cmidrule{2-3}
                                 
                                 & Material                                    & \begin{tabular}[c]{@{}l@{}}can {[}Head{]} be made of {[}Tail{]}? {[}MASK{]}!\\ can {[}Head{]} be made of {[}Tail{]}? {[}MASK{]}.\\ is {[}Head{]} made of {[}Tail{]}? {[}MASK{]}!\\ is {[}Head{]} made of {[}Tail{]}? {[}MASK{]}.\\ is {[}Tail{]} the necessary material for making {[}Head{]}? {[}MASK{]}.\\ is {[}Tail{]} the necessary material for making {[}Head{]}? {[}MASK{]}!\\ does {[}Head{]} consist of {[}Tail{]}? {[}MASK{]}.\\ is {[}Head{]} made up of {[}Tail{]}? {[}MASK{]}.\\ Q: is {[}Head{]} made of {[}Tail{]}? A: {[}MASK{]}. \\ Question: is {[}Head{]} made of {[}Tail{]}? Answer: {[}MASK{]}.\end{tabular}                                                \\
\bottomrule
\end{tabularx}
\end{table*}

\begin{table*}[h!]
\centering
\caption{Prompts for Causal Language Models}
\scriptsize 
\begin{tabularx}{\textwidth}{c|c|l}
\toprule
Model                            & Task                                        & Prompt                                                                                                                                                                                                                                                                                                    \\
\midrule
\multirow{31}{*}{OPT}             & Size, Height, Temperature, Weight, Hardness & \begin{tabular}[c]{@{}l@{}}the {[}Head{]} is {[}Rel{]} than the {[}Tail{]}.\\ {[}Head{]} is {[}Rel{]} than {[}Tail{]}.\\ acutally, the {[}Head{]} is {[}Rel{]} than the {[}Tail{]}.\\ acutally, {[}Head{]} is {[}Rel{]} than {[}Tail{]}.\\ it is well-known that {[}Head{]} is {[}Rel{]} than {[}Tail{]}.\\ {[}Head{]} is indeed {[}Rel{]} than {[}Tail{]}.\\ the {[}Head{]} is indeed {[}Rel{]} than {[}Tail{]}.\\ compared with the {[}Head{]}, the {[}Tail{]} is {[}Rel{]}.\\ a/(an) {[}Head{]} is {[}Rel{]} than a/(an) {[}Tail{]}.\\ yes, {[}Head{]} is {[}Rel{]} than {[}Tail{]}.\end{tabular}                                                                              \\
\cmidrule{2-3}
                                 & Color                                       & \begin{tabular}[c]{@{}l@{}}{[}Head{]} can be of the color {[}Tail{]}.\\ the {[}Head{]} can be of color {[}Tail{]}.\\ the color of a(an) {[}Head{]} is {[}Tail{]}.\\ the color of {[}Head{]} is {[}Tail{]}.\\ the {[}Head{]} is in {[}Tail{]}.\\ {[}Head{]} is {[}Tail{]}.\\ what color is the {[}Head{]}? {[}Tail{]}.\\ {[}Head{]}'s color is {[}Tail{]}.\\ usually, {[}Head{]} is in {[}Tail{]}.\\ {[}Head{]} is typically {[}Tail{]}.\end{tabular}                                                                                                                                                                                                                              \\
                                 \cmidrule{2-3}
                                 & Shape                                       & \begin{tabular}[c]{@{}l@{}}% {[}Head{]} has shape {[}Tail{]}.\\ the {[}Head{]} has shape {[}Tail{]}.\\ {[}Head{]} in the shape of {[}Tail{]}.\\ {[}Head{]} has a shape of {[}Tail{]}.\\ {[}Head{]} has the shape of {[}Tail{]}.\\ the {[}Head{]} has a shape of {[}Tail{]}.\\ 
                                 {[}Head{]} is usually {[}Tail{]}.\\ what is the shape of {[}Head{]}? {[}Tail{]}.\\ {[}Head{]} is typically {[}Tail{]}.\\ {[}Head{]}'s shape is {[}Tail{]}.\end{tabular}                                                                                                                                                                                                                               \\
                                 \cmidrule{2-3}
                                 & Material                                    & \begin{tabular}[c]{@{}l@{}}{[}Head{]} is made of {[}Tail{]}.\\ the {[}Head{]} is made of {[}Tail{]}.\\ {[}Head{]} consists of {[}Tail{]}.\\ the main material of {[}Head{]} is {[}Tail{]}.\\ {[}Tail{]} is necessary material for making {[}Head{]}.\\ the {[}Head{]} consists of {[}Tail{]}.\\ the {[}Head{]} can be made of {[}Tail{]}.\\ the {[}Head{]} is built with {[}Tail{]}.\\ the {[}Head{]} contains {[}Tail{]}.\\ the {[}Head{]} is made up of {[}Tail{]}.\end{tabular}                                                                                                                                              \\
\bottomrule
\end{tabularx}
\end{table*}

\begin{table*}[tbh!]
\centering
\scriptsize 
\caption{Prompts used for VLMs of CLIP.}
\begin{tabularx}{\textwidth}{c|c|l}
\toprule
Model                            & Task                                        & Prompt                                                                                                                                                                                                                                                                                                                                                                                                                                                                                                                                                                                                                                                                                                                    \\
\midrule
CLIP                             & All Tasks                                   & \begin{tabular}[c]{@{}l@{}}a photo of a {[}Head{]}/{[}Attribute{]}.\\ a photo of the {[}Head{]}/{[}Attribute{]}.\\ a blurry photo of a {[}Head{]}/{[}Attribute{]}.\\ a good photo of a {[}Head{]}/{[}Attribute{]}.\\ a painting of a {[}Head{]}/{[}Attribute{]}.\\ a bad photo of a {[}Head{]}/{[}Attribute{]}.\\ a close-up photo of a {[}Head{]}/{[}Attribute{]}.\\ a bright photo of the {[}Head{]}/{[}Attribute{]}.\\ a photo of one {[}Head{]}/{[}Attribute{]}.\\ a low resolution photo of a {[}Head{]}/{[}Attribute{]}.\end{tabular}
\\
\bottomrule
\end{tabularx}
\end{table*}

\end{document}